\documentclass[journal]{IEEEtran}

\usepackage{subfigure}
\usepackage{setspace}

 \usepackage{amsmath,amssymb,amsfonts} 
\usepackage{comment} 
\usepackage{color}
\usepackage{graphicx}
\usepackage[lined,algonl,boxed]{algorithm2e}
\usepackage{url}

\newcommand{\real}{{\mathbb{R}}} \newcommand{\reals}{\real}

\renewcommand{\natural}{{\mathbb{N}}} \newcommand{\naturals}{\natural}

\newcommand{\map}[3]{#1: #2 \rightarrow #3}

\newcommand{\ETSP}{\ensuremath{\operatorname{ETSP}}}

\newcommand{\area}{\operatorname{A}}
\newcommand{\domain}{\mathcal{Q}}
\newcommand{\subd}{\mathcal{S}}
\newcommand{\vdom}{\mathcal{V}}

\newcommand{\E}[1]{\operatorname{E}\left[#1\right]}
\newcommand{\Var}[1]{\operatorname{Var}\left[#1\right]}

\newcommand{\diam}{\operatorname{diam}}

\newtheorem{theorem}{Theorem}
\newtheorem{proposition}[theorem]{Proposition}
\newtheorem{lemma}[theorem]{Lemma}

\newtheorem{example}[theorem]{Example}

\def\QEDclosed{\mbox{\rule[0pt]{1.3ex}{1.3ex}}} 

\def\QED{\QEDclosed} 
\def\proof{\noindent\hspace{2em}{\itshape Proof: }}

\def\endproof{\hspace*{\fill}~\QED\par\endtrivlist\unskip}

\begin{document}

\title{Optimal Foraging of Renewable Resources}

\author{John J. Enright~\thanks{John J. Enright is with Kiva Systems, \texttt{jenright@kivasystems.com}.}
   \, Emilio Frazzoli~\thanks{Emilio Frazzoli is with the Laboratory for
    Information and Decision Systems at the Massachusetts Institute of
    Technology, \texttt{frazzoli@mit.edu}.}}


\maketitle

\begin{abstract}
Consider a team of agents in the plane searching for and visiting target points that appear in a bounded environment according to a stochastic renewal process with a known absolutely continuous spatial distribution. Agents must detect targets with limited-range onboard sensors. It is desired to minimize the expected waiting time between the appearance of a target point, and the instant it is visited. When the sensing radius is small, the system time is dominated by time spent searching, and it is shown that the optimal policy requires the agents to search a region at a relative frequency proportional to the square root of its renewal rate. On the other hand, when targets appear frequently, the system time is dominated by time spent servicing known targets, and it is shown that the optimal policy requires the agents to service a region at a relative frequency proportional to the cube root of its renewal rate. Furthermore, the presented algorithms in this case recover the optimal performance achieved by agents with full information of the environment. Simulation results verify the theoretical performance of the algorithms. 
  \end{abstract}

\section{Introduction}
A  very active research area today addresses the coordination of several mobile agents: groups of autonomous robots and large-scale mobile networks are being considered for a broad class of applications, ranging from environmental monitoring, to search and rescue operations, and national security.  Wide-area surveillance is one of the prototypical missions for Uninhabited
Aerial Vehicles (UAVs): low-altitude UAVs on such a mission must provide coverage of a
region and investigate events of interest as they manifest themselves. In
particular, we are interested in cases in which close-range information is
required on targets detected by onboard sensors, and the UAVs must proceed to the locations to gather further
on-site information.

We address a routing problem for a team of agents in the plane: target
  points appear over time in a bounded environment according to a stochastic renewal process with a known absolutely continuous spatial distribution.  It is desired to stabilize the outstanding target queue and minimize the
  expected elapsed time between the appearance of a target point, and the
  instant it is visited (the system time). This is a
formulation of the Dynamic Traveling Repairman Problem
(DTRP), introduced in~\cite{Psaraftis:88} and thoroughly developed in~\cite{Bertsimas.vanRyzin:91,Bertsimas.vanRyzin:93b}.  Numerous algorithms are presented and analyzed in this series of seminal works.  Furthermore, the property of {\em spatial bias} is studied.  In particular, they analyze the problem and develop policies under the constraint that a target's expected waiting time must be independent of its location.  In addition to combinatorial and convex optimization, many of the solutions rely heavily on results from the relatively mature fields of facility location, probability and queueing theory.  

In an effort to address issues relevant in applications such as autonomous mobile robotics, this paper focuses on a variation of the DTRP.  We place limitations on the information available to the vehicles and analyze the effect on the system's achievable performance.  In particular, we consider the case in which vehicles are not aware of the location of targets as they appear, but rather must detect them using on-board sensors with a limited range. 



The recent literature concerning problems of this class is vast.  Some recent work on the routing of nonholonomic vehicles includes~\cite{Savla.Frazzoli.ea:TAC08,Enright.Savla.Bullo.Frazzoli:09}.  Many of the new results for the small sensing radius case are applicable to coverage
problems~\cite{Enright.Savla.ea:CDC08,Choset:00,Guo.Balakrishnan:06}, in which the agents spread out with some sense of balance, or comb the environment efficiently.  This case also has connections to the Persistent Area Denial (PAD) and area coverage problems~\cite{Schumacher.Chandler.ea:03,Liu.Cruz.ea:04,Cortes.Martinez.ea:04,Moore.Passino:05}. Other works are concerned with the generation of efficient cooperative strategies for several mobile agents to move through a certain number of given target points, possibly avoiding obstacles or threats~\cite{Beard.McLain.ea:02,Earl.DAndrea:05,Li.Cassandras:06,Ma.Castanon:06}.  Trajectory efficiency in these cases is understood in terms of cost for the agents: in other words, efficient trajectories minimize the total path length, the time needed to complete the task, or the fuel/energy expenditure.  A related problem has been investigated as the Weapon-Target Assignment (WTA) problem, in which mobile agents are allowed to team up in order to enhance the probability of a favorable outcome in a target engagement~\cite{Murphey:99,Arslan.Marden.ea:07}. In other works addressing UAV task-assigment, target locations are known and an assignment strategy is sought that maximizes the global success rate~\cite{Bethke.Valenti.ea:08,Alighanbari.Bertuccelli.ea:06} . Moreover, this work holds a place in the recent wave of investigation into the cooperative control of multi-agent systems~\cite{Shamma:07,Zavlanos.Pappas:07}. Other works addressing cooperative task completion by UAVs include~\cite{Ousingsawat.Campbell:07,Bertuccelli.Alighanbari.How:04,Kish.Jacques.ea:07}.


Song and coworkers considered the problem of searching for a static object emitting intermittent stochastic signals under a limited sensing range, and analyze the performance of standard algorithms such as systematic sweep and random walks~\cite{SongKimYi:10}. Due to the intermittent signals from the object, robots must perform a persistent search, thus making the work similar to ours. However, the authors assumed no prior information about the location of the target object is available; hence, their setting is equivalent to the assumption of a uniform spatial distribution. In our work, we explicitly consider non-uniform spatial distributions, which lead to different kinds of optimal policies. Mathew and Mezic presented an algorithm named Spectral Multiscale Coverage (SMC) to devise trajectories such that the spatial distribution of a patrol vehicle's position asymptotically matches a given function~\cite{MathewMezic:09}. Similarly, Cannata and Sgorbissa describe an algorithm that solves what they call the Multirobot Controlled Frequency Coverage (MRCFC) problem, in which a team of robots are required to repeatedly visit a set of predefined locations according to a specified frequency distribution~\cite{Cannata.Sgorbissa:11}. We show that when attempting to minimize discovery time, the desired spatial distribution of the agent's position is dependent on, but not equivalent to the underlying spatial distribution of incidents it must find.

The main contributions of this paper are the following. 1) For the case of small sensing radius, we establish a new lower bound and 2) a new policy whose performance is tight with the lower bound. 3) In combination, these two contributions show that the optimal policy requires that agents search subregions of the environment at relative frequencies proportional to the square root of their renewal rates, i.e., searching is biased towards high density regions, but not as biased as the density function itself. 4) For the case of high target renewal rate, we present a new policy that works for agents with limited sensing capabilities. 5) Comparing the performance of our policy with a previously known lower bound, we show that--like the full-information case--the optimal policy requires that agents search subregions of the environment at relative frequencies proportional to the cube root of their renewal rates. 6) These results imply that the limited sensing capabilities do not adversely affect the optimal performance of the agents in this case, i.e., we recover the optimal performance of the full-information heavy load case. 7) Moreover, we provide scalable, decentralized strategies by which a multi-vehicle team can operate with the above mentioned algorithms, and retain optimal performance. Earlier versions of this work only consider a single agent and uniform spatial distribution~\cite{Enright.Frazzoli:GNC06} or a single agent and a known absolutely continuous spatial distribution, but without analysis of the case of high target renewal rate~\cite{Huynh.Enright.Frazzoli:CDC10}. 

This paper is organized as follows.  In Sec. \ref{sec:setup} we formulate the DTRP with limited sensing and review known results.  In Sections \ref{sec:lb} we offer a lower bound for this new problem. In \ref{sec:alg} we present algorithms for the single agent, and compare their performance with lower bounds. In Sec. \ref{sec:multiple} we adapt our algorithms to the multiple vehicle setting.  Sec. \ref{sec:con} concludes the paper and notes possibilities for future research.  


\section{Problem Formulations and Previous Results}
\label{sec:setup}
In this section, we formally introduce the dynamic vehicle routing problem we wish to study, without the additional limitations on sensing or motion constaints.  We also review results of well studied static vehicle routing problems, in which the vehicles have full information, and travel cost is simply Euclidean distance.  The known performance limits for these problems serve as reference points for results found on the problem variations studied herein.  They give insight as to how the new constraints affect the efficiency of the system.        

Given a set $\mathcal{D}_n \subset \reals^2$ of $n$ points, the two-dimensional Euclidean Traveling Salesman Problem (ETSP) is the problem of finding the shortest tour (closed path) through all points in $\mathcal{D}_n$; let $\ETSP(\mathcal{D}_n)$ be the length of such a tour.  Furthermore, we will make use of the following remarkable result.  
\begin{theorem}[\cite{Beardwood.Halton.ea:59,Steele:90}]
\label{thm:steele}
If the locations of the points in $\mathcal{D}_n$ are independently and identically distributed (i.i.d.) with compact support $\mathcal{Q} \subset \reals^2$, then with probability one  
\begin{equation}
\label{eq:steele}
\lim_{n\rightarrow \infty}\frac{\ETSP(\mathcal{D}_n)}{\sqrt{n}} = \beta \int_{\mathcal{Q}} \sqrt{\varphi(q)} \; dq,
\end{equation}
where $\beta > 0$ is a constant not depending on the distribution of the points and where $\varphi$ is the density of the absolutely continuous part of the distribution of the points.
\end{theorem}
The current best estimate of the constant is $\beta \approx 0.7120$~\cite{Percus.Martin:96,Johnson.McGeoch.ea:96}.  According to~\cite{Larson.Odoni:81}, if $\mathcal{Q}$ is a ``fairly compact and fairly convex'' set in the plane, then \eqref{eq:steele} provides an adequate estimate of the optimal ETSP tour length for values of $n$ as low as $15$.  Interestingly, the asymptotic cost of the ETSP for uniform point distributions is an upper bound on the asymptotic cost for general point distributions, as can be proved by applying Jensen's inequality to \eqref{eq:steele}. In other words, if $\mathrm{A}$ is the area of set $\mathcal{Q}$,
$$\lim_{n\rightarrow \infty}\frac{\ETSP(\mathcal{D}_n)}{\sqrt{n}} = \beta \int_{\mathcal{Q}} \sqrt{\varphi(q)} \; dq \le \beta \sqrt{\mathrm{A}}$$ with probability one.  

We will present algorithms that require online
solutions of large ETSPs. In practice, these solutions are computed using heuristics such as Lin-Kernighan's~\cite{Lin.Kernighan:73} or approximation algorithms such as Christofides'~\cite{Christofides:72}. If the algorithm used in practice guarantees a performance within a constant factor of the optimal,
the effect on the performances of our algorithms
can be modeled as a scaling of the constant
$\beta$.

The following is a formulation of the {\em Dynamic Traveling Repairman Problem} (DTRP)~\cite{Bertsimas.vanRyzin:91,Bertsimas.vanRyzin:93,Bertsimas.vanRyzin:93b}.  Let $\mathcal{Q} \subset \reals^2$ be a convex, compact 
domain on the plane, with non-empty interior; we will refer to $\mathcal{Q}$ as the {\em environment}. Let $A$ be the area of $\mathcal{Q}$. {\em Target} points are i.i.d. and generated according to a spatio-temporal Poisson point process, with temporal intensity $\lambda >0$, 
and an absolutely continuous spatial distribution described by the density function $\varphi: \mathcal{Q} \rightarrow \reals_+$. The spatial density function $\varphi$ is K-Lipschitz, $\left| \varphi(q_1) - \varphi(q_2) \right| \le K \|q_1-q_2 \|$ for all $q_1$ and $q_2$ in $\mathcal{Q}$, and bounded above and below, $0<\underline \varphi \le \varphi(q) \le \overline \varphi < \infty$ for all $q$ in $\mathcal{Q}$, and is normalized in such a way that $\int_\mathcal{Q} \varphi(q) \; dq  = 1$.
For any $t > 0$,  $\mathcal{P}(t)$ is a random collection of points in $\mathcal{Q}$, representing the targets generated in the time interval $[0, t)$.  The following are consequences of the properties defining Poisson processes. 
\begin{itemize}
\item The total numbers of targets generated in two disjoint time-space regions are {\em independent} random variables.
\item The total number of targets generated in an interval $[t,t+\Delta t)$  in a measurable set $\mathcal{S} \subseteq \mathcal{Q}$
satisfies 
\begin{align*}
\label{eqn:poisson} 
\mathrm{ Pr}\left[\mathrm{card}\left( (\mathcal{P}(t+\Delta t) - \mathcal{P}(t)) \cap \mathcal{S} \right) =k \right] \\
=\frac{ \exp(-\lambda \Delta t \cdot \varphi(\mathcal{S}))(\lambda  \Delta t\cdot \varphi(\mathcal{S}))^k}{k!}, 
\end{align*}
for all $k \in \naturals$, and hence
$$ \mathrm{E}[\mathrm{card}\left( (\mathcal{P}(t+\Delta t) - \mathcal{P}(t)) \cap \mathcal{S}\right)]=
\lambda \Delta t \cdot \varphi(\mathcal{S}),$$
where $\varphi(\mathcal{S})$ is a shorthand for $\int_\mathcal{S} \varphi(q) \; dq$.
\end{itemize}


A service request is fulfilled when one of $m$ mobile agents, modeled as point masses, moves to the target point associated with it; $m$ is a possibly large, but finite number. In later sections, we will introduce limitations on the agent's awareness of targets, and nonholonomic constraints on the agent's motion. Let us assume the agents have holonomic (single integrator) dynamics, and that all agent's are aware of a target's location upon its arrival epoch. Let 
$p(t) = \left(p_1(t), p_2 (t), \ldots, p_m(t)\right) \in \mathcal{Q}^m$ be a vector describing the positions of the agents at time $t$.  The agents are free to move, with bounded speed, within the  workspace $\mathcal{Q}$; let $v$ be the maximum speed of the agents. 
In other words,  the dynamics of the agents are described by differential equations of the form 
\begin{equation*}
\label{eq:model}
\frac{d\; p_i(t)}{dt} = u_i(t), \quad \mbox{ with } \|u_i(t)\| \le v,\quad \forall t \ge 0, i \in I_m,
\end{equation*} 
where we denote $I_m = \{1, \ldots, m\}$.  The agents are identical, and have unlimited fuel and target-servicing capacity. 

 
Let $\mathcal{D}: t \rightarrow 2^{\mathcal{Q}}$ indicate (a realization of) the stochastic process obtained combining the target generation process $\mathcal{P}$ and the  removal process caused by the agents visiting outstanding requests.
The random set $\mathcal{D}(t) \subset \mathcal{Q}$ represents the {\em demand}, i.e., the service requests outstanding at time $t$; let $n(t) = \mathrm{card}(\mathcal{D}(t))$.


A motion coordination policy is a function that determines the actions of each vehicle over time, based on the locally-available information. For the time being, we will denote these functions as $\pi = (\pi_1, \pi_2, \ldots, \pi_m)$, but do not explicitly state their domain; the output of these functions is a velocity command for each agent. Our objective is the design of motion coordination strategies that allow the mobile agents to fulfill service requests efficiently (we will make this more precise in the following). 

A policy $\pi = (\pi_1, \pi_2, \ldots, \pi_m)$ is said to be {\em stabilizing} if, under its effect, the expected number of outstanding targets does not diverge over time, i.e., if 
\begin{equation*}
\overline n_\pi = \lim_{t\rightarrow \infty} \mathrm{E}\left[ n(t) : \mbox{ agents execute policy } \pi \right] < \infty.
\end{equation*}
Intuitively, a policy is stabilizing if the mobile agents are able to visit targets at a rate that is---on average---at least as fast as the rate at which new service requests are generated. 

Let $T_j$ be the time elapsed between the generation of the $j$-th target, and the time it is fulfilled. If the system is stable, 
then the following balance equation (also known as Little's formula~\cite{Little:61}) holds:
\begin{equation}
\label{eq:little}
\overline n_\pi  = \lambda \overline T_\pi,
\end{equation}
where $\overline{T}_\pi:=\lim_{j\rightarrow \infty} \mathrm{E}[T_j]$ is the system time under policy $\pi$, i.e., the expected time a target must wait before being fulfilled, given that  the mobile agents follow the strategy defined by $\pi$.  Note that  the system time $\overline T_\pi$ can be thought of as a measure of the quality of service collectively provided by the agents. 

At this point we can finally state our problem: we wish to devise a policy that is (i) stabilizing, and (ii) yields a quality of service (system time)  achieving, or approximating, the theoretical optimal performance given by 
\begin{equation*}
\label{eq:opt}
\overline T_\mathrm{opt} = \inf_{\pi \text{ stabilizing}} \overline T_\pi.
\end{equation*}

In the following, we are interested in designing control policies that
provide constant-factor approximations of the optimal achievable
performance; a policy $\pi$ is said to provide a constant-factor
approximation of $\kappa$ if $\overline{T}_\pi \le \kappa \overline{T}_\mathrm{opt}$.  
Furthermore, a policy is called {\em spatially unbiased} if, under its action, a target's expected waiting time is independent of its location and {\em spatially biased} otherwise.  We shall investigate how this spatial constraint effects the achievable system time, i.e., we shall find lower bounds and develop algorithms within the class of spatially unbiased policies, and without.    
Moreover, we are interested in decentralized, scalable, adaptive control policies, that
rely only on local exchange of information between neighboring
vehicles, and do not require explicit knowledge of the global
structure of the network. 

The DTRP with general demand distribution is studied in~\cite{Bertsimas.vanRyzin:93b}, where the form of the optimal system time in heavy load is first derived.  However, there remained a constant-factor gap between the lower and upper bounds.  The coefficient of the lower bound was tightened from $\left(2/(3\sqrt{\pi})\right)^2$ to $\beta^2/2$ in~\cite{Xu:94}:
\begin{equation}
\label{eq:bertsimas}
\lim_{\lambda \rightarrow \infty} \frac{\overline T_\mathrm{opt}}{\lambda} = \frac{\beta^2}{2m^2v^2} \left( \int_\mathcal{Q} \sqrt{\varphi(q)} \ dq \right)^2,
\end{equation}
within the class of spatially unbiased policies, and
\begin{equation}
\label{eq:bertsimas2}
\lim_{\lambda \rightarrow \infty} \frac{\overline T_\mathrm{opt}}{\lambda} = \frac{\beta^2}{2m^2v^2} \left( \int_\mathcal{Q} \varphi(q)^{2/3} \ dq \right)^3,
\end{equation}
within the class of spatially biased policies.  As mentioned in~\cite{Bertsimas.vanRyzin:93b},
$$\mathrm{A}\ge \left( \int_\mathcal{Q} \sqrt{\varphi(q)} \ dq \right)^2\ge \left( \int_\mathcal{Q} \varphi(q)^{2/3} \ dq \right)^3$$
with equality holding throughout if and only if $\varphi(q) = 1/\mathrm{A}$ for all $q\in \mathcal{Q}$.  In other words, uniform density is the worst possible, and any non-uniformity will strictly lower the optimal system time.  This is analogous with the length of the stochastic ETSP, i.e., Eq. \eqref{eq:steele}.  Furthermore, the optimal system time for spatially biased policies is lower than that of spatially unbiased policies.  This follows intuition as spatially unbiased waiting time is a constraint limiting the realm of available policies.

In addition to the above formulation of the DTRP, we add a constraint on the information available to the agents.  Agents are not aware of a target's existence or location upon its arrival epoch.  Rather, they must detect targets with limited-range onboard sensors, i.e., they must come within the local vicinity of the target.  Let us call this variation the Limited Sensing DTRP.  Formally, this means the set $\mathcal{D}(t)$ is in general not entirely known to all agents, due to the fact that the sensing range is limited. For the sake of simplicity, we will model the sensing region of an agent as a disk of radius $r$ centered at the agent's position; indicate with 
$$d(p_i) = \{ q \in \reals^2: \|p_i-q\| \le r \}$$
the sensing region for the $i$-th agent. Other shapes of the sensor footprint can be considered with minor modifications to our analysis, and affect the results at most by a constant. We will assume that the sensor footprint is small enough that it is contained within $\mathcal{Q}$ for at least one position of the agent; moreover, we will be interested in analyzing the effect of a limited sensor range as $r \to 0^+$.  

\section{Lower Bounds on the Optimal System Time}
\label{sec:lb}
Every target must be detected by an agent's sensor before it is serviced. Let $T_j^\mathrm{detect}$ be the time elapsed between the generation of the $j$-th target, and the time it is first detected by any agent's sensor. Consider an alternative scenario we call the {\em Detection DTRP}, in which targets are fulfilled at the instant they are first detected, and $\overline{T}^\mathrm{detect}_\pi:=\lim_{j\rightarrow \infty} \mathrm{E}[T_j^\mathrm{detect}]$ is the {\em detection time} under policy $\pi$.  For all targets, $T_j^\mathrm{detect} \le T_j$, and thus    
\begin{equation*}
\label{eq:infdetect}
 \inf_{\pi \text{ stabilizing}} \overline T_\pi \ge \inf_{\pi \text{ stabilizing}} \overline T^\mathrm{detect}_\pi.
\end{equation*}
In other words, a lower bound on the achievable detection time is also a lower bound on the achievable system time, $\overline{T}_\mathrm{opt}$. We leverage this fact in the following theorem. 

\begin{theorem}
\label{thm:sslb}
The optimal system time for the DTRP with limited sensing satisfies
\begin{equation}
\label{eq:ssulb}
\lim_{r \rightarrow 0^+ }\overline T_\mathrm{opt} r \ge \frac{\mathrm{A}}{4mv},
\end{equation}
  within the class of spatially unbiased policies and
  \begin{equation}
\label{eq:ssblb}
\lim_{r \rightarrow 0^+ }\overline T_\mathrm{opt} r \ge \frac{1}{4mv}  \left( \int_\mathcal{Q} \sqrt{\varphi(q)} \ dq \right)^2,
\end{equation}
 within the class of spatially biased policies.
\end{theorem}
\proof 
The probability that a target's location is within a sensor footprint at the time of arrival is bounded by
$$\mathrm{Pr}[q \in \cup_{i=1}^m d(p_i)] \le \overline \varphi m \pi r^2.$$
In this case, $T_j^\mathrm{detect} = 0$.  However, for any fixed number of vehicles $m$ and distribution $\varphi$,
$$\lim_{r \rightarrow 0^+} \mathrm{Pr}[q \in \cup_{i=1}^m d(p_i)] \le \lim_{r \rightarrow 0^+}\overline \varphi m \pi r^2 = 0,$$
and therefore,
$$\lim_{r \rightarrow 0^+} \mathrm{Pr}[q \notin \cup_{i=1}^m d(p_i)] = 1.$$
In this limit, from the perspective of a point $q \in \mathcal{Q}$, the actions of a given stabilizing policy $\pi$ are described by the following (possibly random) sequence of variables: 
the lengths of the time intervals during which the point is not contained in any sensor footprint, $Y_k(q)$.  In the following, we use the notation 
$$\E{Y_\pi (q)} = \lim_{k \rightarrow \infty }\E{Y_k(q): \mbox{ agents } \mbox{ execute policy } \pi}.$$ 
Due to random incidence~\cite{Larson.Odoni:81,Gallager:96}, a target's detection time, conditioned upon its location, can be written as
\begin{align*} \lim_{r \rightarrow 0^+}\E{T^\mathrm{detect} | q} &= \lim_{r \rightarrow 0^+} \mathrm{Pr}[q \notin \cup_{i=1}^m d(p_i)]\cdot \frac{\E{Y^2_\pi(q)}}{2 \E{Y_\pi(q)}} \\
&=   \frac{\E{Y_\pi(q)}^2 + \Var{Y_\pi(q)}}{2 \E{Y_\pi(q)}} \\
&\ge \frac{1}{2} \E{Y_\pi (q)}\\
\end{align*}
where $\E{Y^2_\pi(q)}$ and $\Var{Y_\pi(q)}$ are, respectively, the second moment and variance of the sequence $Y_k(q)$ under the actions of policy $\pi$.  In other words, for fixed 
$\E{Y_\pi (q)}$, the system time is minimized if $\Var{Y_\pi(q)} = 0$.  This occurs under the actions of a deterministic policy with exactly regular time intervals between searching location $q$.  

Define $f(q)$ as the---time averaged---frequency at which point $q$ is searched under the actions of a policy.  Note that $f(q) = 1/\E{Y_\pi (q)}$, and so     
\begin{align*}
\overline T^\mathrm{detect} &= \int_\mathcal{Q} \varphi(q) \E{T_j^\mathrm{detect} | q} \ dq \\
&\ge  \frac{1}{2} \int_\mathcal{Q} \frac{ \varphi(q)}{f(q)} \ dq. \\
\end{align*}
The $m$-vehicle system is capable of searching at a maximum rate of $2mvr$ (area per unit time), and so the average searching frequency is bounded by $\int_\mathcal{Q}  f(q)/A \ dq \le 2mvr/\mathrm{A}$. Recalling $\overline T_\mathrm{opt} \ge \overline T^\mathrm{detect}$ we have the following minimization problem:
\begin{align*}
2 \overline T_\mathrm{opt} &\ge \min_f  \int_\mathcal{Q} \frac{\varphi(q)}{f(q)} \ dq \ \ \ \ \mbox{subject to}  \\  &\int_\mathcal{Q} f(q) \ dq \le 2mvr \ \ \ \ \mbox{and} \ \ \ \ f(q) > 0.
\end{align*}
Since the objective function is convex in $f(q)$ and the constraints are linear, the above is an infinite-dimensional convex program.  Relaxing the constraint with a multiplier, we arrive at the Lagrange dual: 
\begin{align*}
\label{eqn:freqlagrange}
2 \overline T_\mathrm{opt} &\ge \min_{f(q)>0}\left[ \int_\mathcal{Q} \frac{\varphi(q)}{f(q)} \ dq + \Gamma \left(  \int_\mathcal{Q} f(q) \ dq - 2mvr \right) \right]\\
&=  \int_\mathcal{Q} \min_{f(q)>0}\left[ \frac{\varphi(q)}{f(q)} + \Gamma f(q) \right] \ dq  - 2mvr\Gamma.
\end{align*}
Differentiating the integrand with respect to $f(q)$ and setting it equal to zero, we find the pair 
\begin{equation}
\label{eq:convexsoln}
f^*(q) = \sqrt{\frac{\varphi(q)}{\Gamma^*}}
\end{equation}
and $$\Gamma^* = \left( \frac{1}{2mvr}  \int_\mathcal{Q} \sqrt{f^*(q)} \ dq \right)^2$$ satisfy the Kuhn-Tucker necessary conditions for optimality~\cite{Bertsekas:95}, and since it is a convex program, these conditions are sufficient to insure global optimality.  Upon substitution, \eqref{eq:ssblb} is proved.  
On the other hand, the constraint of unbiased service requires that
\begin{align*}
\E{T_j^\mathrm{detect} | q_j = q}  =\frac{1}{2f(q)} = \overline T^\mathrm{detect}, \ \ \forall q \in \mathcal{Q}. \\
\end{align*}
Substituting into $\int_\mathcal{Q} f(q) \ dq \le 2mvr$, we get
$$ \frac{\mathrm{A}}{2 \overline T^\mathrm{detect}} \le 2mvr.$$
Rearranging and recalling $\overline T_\mathrm{opt} \ge \overline T^\mathrm{detect}$, we arrive at \eqref{eq:ssulb}.
\endproof

Oftentimes, a tight lower bound offers insight into the optimal solution of a problem.  Assuming that this lower bound is tight, Eq. \eqref{eq:convexsoln} suggests that in the spatially biased small sensing-range case, the optimal policy searches a point $q$ at regular intervals at a relative frequency proportional to $\sqrt{\varphi(q)}$.  Moreover, we have presented new lower bounds on the optimal system time related to the searching capability of the agents and the necessary struture of any stabilizing policy.  In the following sections, we will use these bounds to evaluate the performance of our proposed policies for the case of small sensor range.

\section{Algorithms and Policies for a Single Agent}
\label{sec:alg}
In this section, we present four policies for the single-vehicle Limited Sensor DTRP, and prove their respective optimality in different limiting cases and classes of the problem, namely, the case of small sensor range (spatially unbiased and biased), and the case of heavy load (spatially unbiased and biased).  To begin, we present two algorithms (subroutines used by the policies) by which an agent can service targets in a given region of the environment.  The first is designed for the small-sensor case, and the second is designed for the heavy load case.  We analyze their properties in their respective cases.  

In the following, we consider a convex subregion $\subd \subseteq \domain$ of area $\area_\subd$.  The targets in $\subd$ are generated by a local Poisson process with time-intensity $\lambda_\subd = \lambda \varphi(\subd)$ and spatial distribution $\varphi_\subd (q) = \varphi(q)/\varphi(\subd)$ for all $q \in \subd$.  Note that $\varphi_\subd$ is  normalized such that $ \varphi_\subd(\subd)  = 1$.  The first algorithm is designed for the case of small sensing-range.    

\subsection*{SWEEP-SERVICE}   
The description of this algorithm requires the use of an inertial Cartesian coordinate frame.  The algorithm is defined as follows.
\begin{itemize}
\item Partition $\subd$ into elements of width $2r$ with lines parallel to the $x$-axis.  
\item Define a strip as the bounding rectangle of an element of the partition, with sides parallel to the coordinate axes, and minimum side-lengths.  
\item Plan a path running along the longitudinal bisector of
each strip, visiting all strips from top-to-bottom, connecting adjacent strip bisectors by their endpoints.   
\item Execute this path and visit targets as they are detected in the following manner.  If a target is detected in the current strip and it is in front of the agent (with respect to the direction the agent is moving on the path), then the agent continues on the path until its position has the same $x$-coordinate as the target's.  It then departs from the path, moving directly to the target, returning to the point of departure, and continuing on the path.  If a target is detected outside the current strip, or behind the agent, then it is ignored.  
\end{itemize}
We now analyze the length $L_\mathrm{swp}(\subd)$ of the path planned by the algorithm.
\begin{proposition}
\label{thm:sweep}
The length $L_\mathrm{swp}(\subd)$ of the path planned by SWEEP-SERVICE for region $\subd$ satisfies
\begin{equation*}
\label{eq:sweep1}
\lim_{r \rightarrow 0^+} L_\mathrm{swp}(\subd) r \le \frac{\area_\subd}{2}
\end{equation*}
\end{proposition}
\proof 
Consider a grid of squares with sides of length $2r$ parallel to the coordinate axes.  Denote $N_\mathrm{sq}(\subd)$ as the number of squares  with nonzero intersection with $\subd$.  The sum of the lengths of all strips can be bounded by $ 2rN_\mathrm{sq}(\subd).$  Bound the region $\subd$ with a rectangle whose sides are parallel with the coordinate axes, and denote the length of its perimeter $P(\subd)$.  The length of the path between the endpoints of the strip bisectors can be bounded above by $P(\subd)$.  Thus, the total path length satisfies
$$L_\mathrm{swp}(\subd) \le 2rN_\mathrm{sq}(\subd)+P(\subd).$$
But the number of squares satisfies~\cite{Kershner:39}
 $$\lim_{r \rightarrow 0^+}N_\mathrm{sq}(\subd) r^2 = \frac{\area_\subd}{4}$$
and so
 $$\lim_{r \rightarrow 0^+}  L_\mathrm{swp}(\subd) r \le  \lim_{r \rightarrow 0^+} (2N_\mathrm{sq}(\subd)r^2+ P(\subd)r ) = \frac{\area_\subd}{2}.$$
\endproof

\begin{figure}
\label{fig:sweepservice}
\centering
\includegraphics[scale=0.25]{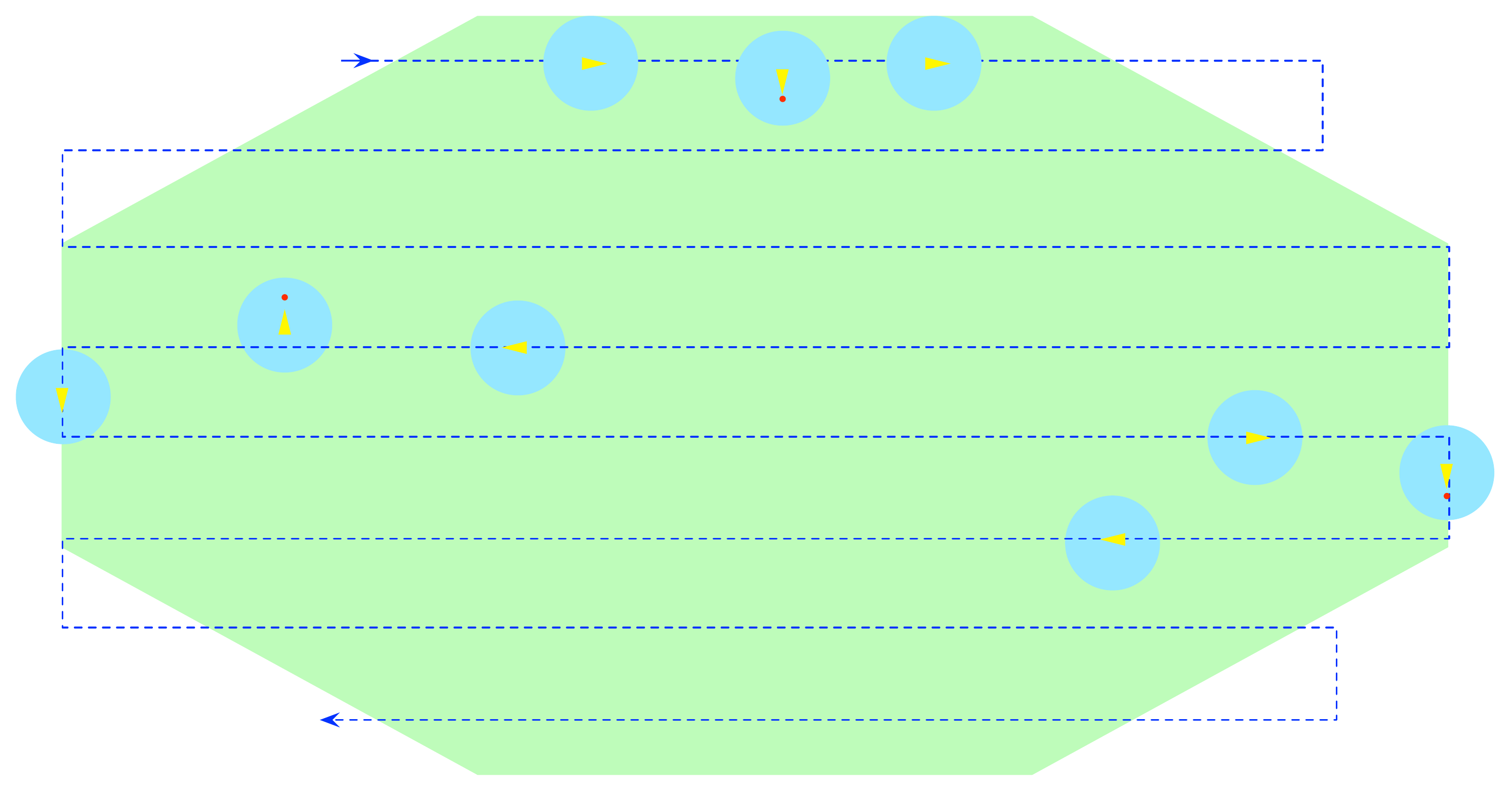}
 \caption{Depiction of an agent executing SWEEP-SERVICE.}
\end{figure}

\subsection*{SNAPSHOT-TSP}  
The second algorithm is designed for the heavy load case.  This algorithm requires that the subregion $\subd \subseteq \domain$ has a size and shape such that it can be contained in the sensor footprint of the agent, i.e., there exists a position $p$ such that $\|p-q\| \le r$ for all $q \in \subd$.  Let $p_\mathrm{snap}$ be one such position.  The algorithm is defined as follows.  
\begin{itemize}
\item Move to location $p_\mathrm{snap}$ and take a snapshot, i.e., store in memory, the locations of all targets outstanding at the current time, called $t_\mathrm{snap}$. 
\item Compute a minimum-length tour of all points in the snapshot. 
\item Generate a uniformly distributed (in terms of path length) random starting position on the tour (not necessarily a target's location).
\item Randomly select a direction (clockwise or counterclockwise) with equal probability.
\item Move to the starting position and execute the tour in the chosen direction, ignoring all targets that appear after $t_\mathrm{snap}$.    
\end{itemize}
We now analyze the length of the tour computed by the algorithm.  Define the set of targets in the snapshot as $\mathcal{D}_\mathrm{snap}$ and denote the cardinality of this set as $n_\mathrm{snap}$.  
\begin{proposition}
\label{thm:snap}
Assuming that all targets generated before some past time $t_\mathrm{clear}$ were cleared from $\subd$, and the set of targets outstanding at the current time $ t$ is the set generated by the local Poisson process in the time interval $(t_\mathrm{clear} , t]$ of length $\Delta t = t-t_\mathrm{clear}$, the length of the tour computed by SNAPSHOT-TSP satisfies
\begin{equation*}
\label{eq:snap5}
\lim_{\lambda \rightarrow \infty} \frac{\E{\ETSP(\mathcal{D}_\mathrm{snap})}}{\sqrt{\lambda}} = \beta \sqrt{\Delta t}  \int_{\subd}\sqrt{ \varphi (q)} \; dq.
\end{equation*}
\end{proposition}
\proof The points in the set $\mathcal{D}_\mathrm{snap}$ are i.i.d. with compact support $\subd \subset \reals^2$, and $n_\mathrm{snap} \rightarrow \infty$ almost surely as $\lambda \rightarrow \infty$.  By Theorem \ref{thm:steele} 
\begin{equation*}
\label{eq:snap1}
\lim_{n_\mathrm{snap}\rightarrow \infty}\frac{\ETSP(\mathcal{D}_\mathrm{snap})}{\sqrt{n_\mathrm{snap}}} = \beta \int_{\subd}\sqrt{ \varphi_\subd(q)} \; dq.
\end{equation*}
Thus as $n_\mathrm{snap}\rightarrow \infty$,
\begin{align*}
\label{eq:snap2}
\E{\ETSP(\mathcal{D}_\mathrm{snap})} &=\E{ \beta \sqrt{n_\mathrm{snap}} \int_{\subd}\sqrt{ \varphi_\subd(q)} \; dq} \\
&=  \beta  \E{\sqrt{n_\mathrm{snap}}}  \int_{\subd}\sqrt{ \varphi_\subd(q)} \; dq. 
\end{align*}
Define the length of the time interval over which $\mathcal{D}_\mathrm{snap}$ was generated as $\Delta t = t_\mathrm{snap} - t_\mathrm{clear}$.  The random variable $n_\mathrm{snap}$ is Poisson with mean $\E{ n_\mathrm{snap}} =\lambda_\subd \Delta t$.  By Jensen's inequality,
$$\E{\sqrt{n_\mathrm{snap}}} \le \sqrt{\E{ n_\mathrm{snap}}}.$$
However, as $\lambda_\subd  \rightarrow \infty$, the above inequality approaches equality, i.e., 
$$\lim_{\lambda_\subd \rightarrow \infty}\frac{\E{\sqrt{n_\mathrm{snap}}}}{\sqrt{\lambda_\subd}} = \sqrt{\Delta t}$$
and thus
\begin{equation*}
\label{eq:snap4}
\lim_{\lambda_\subd \rightarrow \infty} \frac{\E{\ETSP(\mathcal{D}_\mathrm{snap})}}{\sqrt{\lambda_\subd}} = \beta \sqrt{\Delta t}  \int_{\subd}\sqrt{ \varphi_\subd(q)} \; dq.
\end{equation*}
Substituting $\lambda_\subd = \lambda \varphi(\subd)$ and $\varphi_\subd (q) = \varphi(q)/\varphi(\subd)$, we arrive at the claim. \endproof  

The reader might find the choice of taking a snapshot at a particular instant and ignoring all targets generated thereafter peculiar.  One might suggest that the agent could easily service newly generated targets whose locations happen to coincide with the vicinity of targets already in the current snapshot.  However, with the described method, the sets of points in the snapshot are i.i.d. from the given Poisson process, and this allows us to apply Theorem \ref{thm:steele} to the tour computed for each snapshot.  

We now present four policies, each of which is designed for one of the four cases: small sensing-range (spatially biased and unbiased) and heavy load (spatially biased and unbiased).  Some of the policies we present require a partition of the environment $\domain$ into tiles $(\subd_1,\subd_2,...,\subd_K)$, i.e., $\cup_{k=1}^K \subd_k = \domain$ and $\subd_k \cap \subd_\ell = \emptyset$ if $k \ne \ell$.  Each policy has different required properties of the partition, however, let us define some of the properties here.  An equitable partition with respect to a measure $\map{\psi}{\domain}{\reals_+}$ is a partition such that $\psi(\subd_k) = \psi(\subd_\ell)$ for all $k,\ell \in I_K.$  Note that this condition implies $\psi(\subd_k) = \psi(\domain)/K$ for all $k \in I_K.$  A convex partition is one whose subsets are convex.  

\subsection{The Unbiased Region Sweep (URS) Policy}  
The policy is defined in Algorithm \ref{alg:urs}.  The index $i$ is a label for the current phase of the policy.
\begin{algorithm} 
\caption{URS Policy} 
\BlankLine 
\For{$i\leftarrow 1$ \KwTo $\infty$}{ 
\BlankLine 
Execute SWEEP-SERVICE on the environment $\domain$
\BlankLine 
} 
\label{alg:urs} 
\end{algorithm}

\begin{theorem}
\label{thm:urs}
Let $\overline T_\mathrm{opt}$ be the optimal system time for the single-agent Limited Sensing DTRP over the class of spatially unbiased policies.  Then the system time of a single agent operating on $\domain$ under the URS policy satisfies
\begin{equation}
\frac{\overline T_\mathrm{URS}}{\overline T_\mathrm{opt}} = 1 \ \ \ \ \mbox{as } r \rightarrow 0^+.
\end{equation}
\end{theorem}
\proof Define a phase of this policy as the time interval over which the agent performs one execution of SWEEP-SERVICE on the region $\domain$.  Denote the length of the $i$-th phase by $T^\mathrm{phase}_i$, and the number of targets visited during the $i$-th phase by $n_i$.  Assuming that the policy is stabilizing, $n_i$ is finite.  The expected length of the $i$-th phase $T^\mathrm{phase}_{i}$, conditioned on $n_{i}$, satisfies
$$ \E{T^\mathrm{phase}_{i} | n_{i}} \le \frac{L_\mathrm{swp}(\domain)+ 2rn_{i}+\diam(\domain)}{v}.$$ 
Applying Proposition \ref{thm:sweep},
\begin{align*}
\lim_{r\rightarrow 0^+} \E{T^\mathrm{phase}_{i} | n_{i}} r &\le \\
\lim_{r\rightarrow 0^+} & \frac{L_\mathrm{swp}(\domain)r + 2n_{i}r^2+\diam(\domain)r}{v} = \frac{\area}{2v}.
\end{align*}
In other words, in the limit as $r \rightarrow 0^+$, the length of the phase does not depend on the number of targets serviced, so long as it is finite. In expectation, a target waits one half of a phase to be serviced, independent of its location. Therefore
$$ \lim_{r\rightarrow 0^+} \overline T_\mathrm{URS}r \le \frac{\area}{4v},$$
and moreover, the policy is spatially unbiased.  Combining the above result with the lower bound on the optimal system time within the class of spatially unbiased policies in Theorem \ref{thm:sslb} and substituting $m=1$, the claim is proved. 
\endproof

\begin{figure}
\label{fig:simurs}
\centering
\includegraphics[scale=0.37]{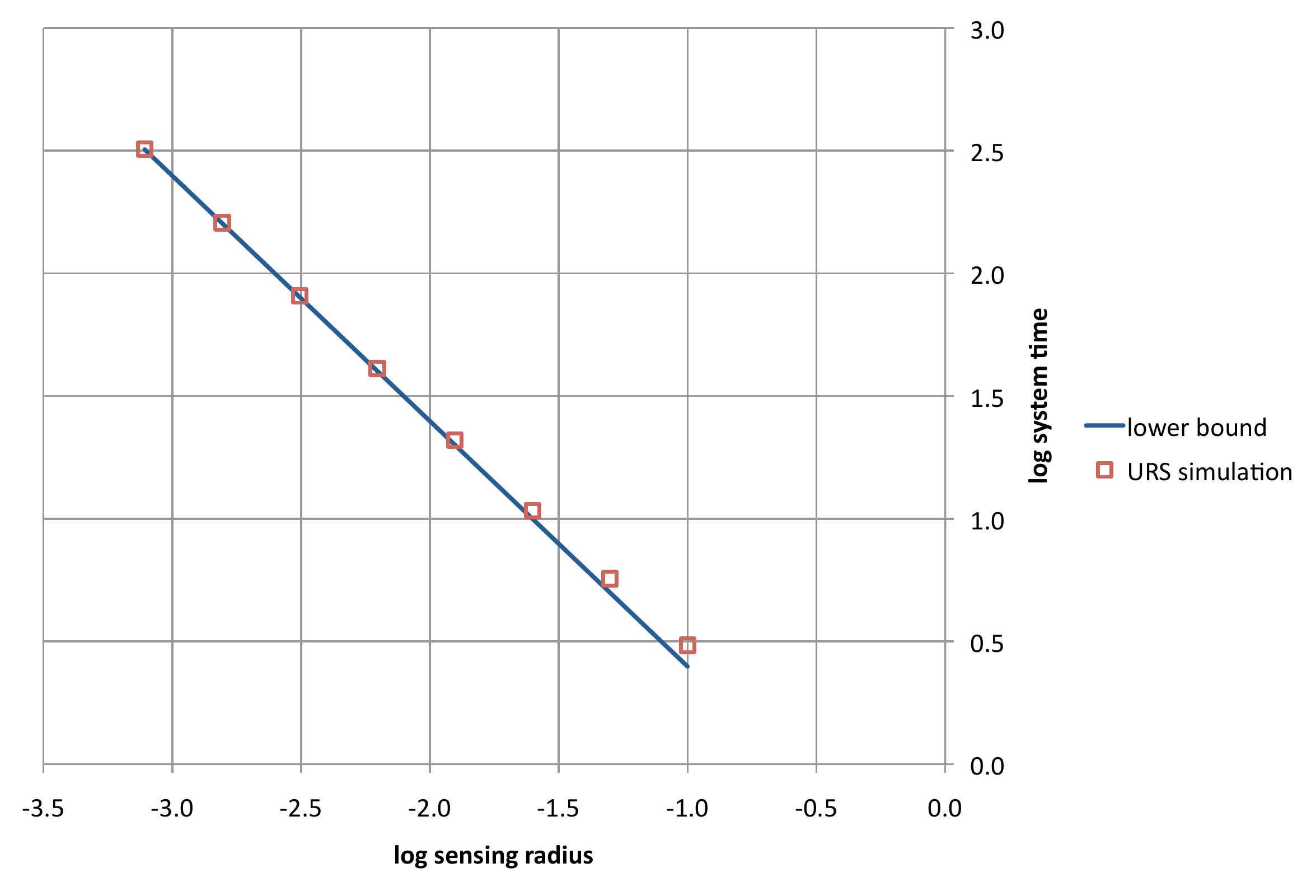}
 \caption{Simulation performance of the URS policy for a single agent with unit velocity in a unit square environment and uniform spatial distribution. Results are compared with the theoretical lower bound (Eq. \eqref{eq:ssulb}) as radius $r$ approaches zero on a log-log plot. The URS policy provides near-optimal performance for very small values of $r$.}
\end{figure}

Theorem \ref{thm:urs} shows that the optimal spatially unbiased policy for small sensing range simply searches the entire environment with equal frequency.  In this case, the constraint on spatial bias does not allow the agent to leverage non-uniformity in the distribution of targets, $\varphi$, in order to lower the system time. We performed numerical experiments of the URS policy and results are shown in comparison with the lower bound (Eq. \eqref{eq:ssulb}) in Fig. \ref{fig:simurs} on a log-log plot. The URS policy provides near-optimal performance for very small values of sensing radius $r$.

\subsection{The Biased Tile Sweep (BTS) Policy}
This policy requires that $\varphi$ be a piecewise uniform density, i.e., $\domain_1,\domain_2,\dots,\domain_J$ be a partition of $\domain$ such that $\varphi(q) = \mu_j \ \forall q \in \domain_j, j=1,2,\dots,J$.  Let us assume that each subset $\domain_j$ is convex, as a non-convex subset can be further partitioned into convex subsets.  Let us denote $\area_j = \mathrm{Area}(\domain_j), \ j=1,2,\dots,J.$  

This policy requires a tiling of the environment $\domain$ with the following properties.  For some positive integer $K \in \naturals$, partition each subset $\domain_j$ into $K_j = K/\sqrt{\mu_j}$ convex tiles, each of area $\area_j/K_j = \area_j\sqrt{\mu_j}/K.$  We assume $K$ is chosen large enough that an integer $K_j$ can be found such that $K/K_j$ is sufficiently close to $\sqrt{\mu_j}.$  In other words, it requires a convex and equitable partition of each region $\domain_j$ with respect to a constant measure $\psi$.  This can be done with the following simple method.  Partition $\domain_j$ into strip-like tiles of equal measure with $K_j$ parallel lines.  Let us give the tiles of $\domain_j$ an ordered labeling $\subd_{j,1},\subd_{j,2},\dots,\subd_{j,K_j}$.  The BTS policy is defined in Algorithm \ref{alg:bts}.  The index $i$ is a label for the current phase of the policy.    

\begin{algorithm} 
\caption{BTS Policy} 
Initialize $k_j \leftarrow 1$ for $j = 1,2,\dots,J$\\
\For{$i\leftarrow 1$ \KwTo $\infty$}{ 
\For{$j\leftarrow 1$ \KwTo $J$}{
Execute SWEEP-SERVICE on tile $\subd_{j,k_j}$\\
\lIf{$k_j < K_j$}{$k_j \leftarrow (k_j+1)$}\\
\lElse{$k_j \leftarrow 1$}
} 
} 
\label{alg:bts} 
\end{algorithm} 

\begin{example}
\label{ex:bts}
In order to illustrate the application of the BTS policy on a specific problem instance, we consider the environment and piece-wise constant density function shown in Fig. \ref{fig:btsexample}. This example is made up of four subregions, $\domain_j$, each of constant density, $\mu_j$, indicated in the drawing. We do not specify the areas of the subregions as their magnitudes are not relevant to the algorithm. For the example to be well posed, we can assume that the areas are such that $\varphi(\domain) = 1$. In fact, the absolute values of the $\mu_j$'s are not relevant either. In order to apply the BTS policy, all we need to know are the relative magnitudes of the $\mu_j$'s. In other words, how much \emph{more} likely is it that a target appears in $\domain_1$ rather than $\domain_2$? Towards the end of making the relative magnitudes clear, this example's lowest density subregion has a density of 1. Any given piecewise-constant density can be normalized to such a form. 

The first step of the algorithm is to choose our scaling constant, a positive integer $K \in \naturals$. Since the highest density is $\mu_1 = 36$ and $K_1 = K/\sqrt{\mu_1}$ in order to ensure that $K_1 \ge 1$, we must set $K \ge 6$. The beneficial aspect of setting $K$ arbitrarily large is that it allows us to make $K/K_j$ arbitrarily close to $\sqrt{\mu_j}$ for each $j$. But in practice, there is a cost associated with increasing $K$: it increases the frequency of transition moves between sweeping tiles. In the limit as $ r \rightarrow 0^+$ these transition costs are negligible as the time required to sweep an individual tile dwarfs them, but for small but fixed $r$, these transition costs must be balanced with the benefit of designing the ideal ratio's between the $K_j$'s. In this particular example, we don't have to make a trade-off between the transition costs and the ideal ratio's of the $K_j$'s. We can set $K$ to its minimum feasible value, $K=6$, and then $K_j =  6/\sqrt{\mu_j}$ for each $j$, resulting in

\begin{align*}
\label{eq:snap2}
K_1 &= 1, \\
K_2 &= 2, \\
K_3 &= 3, \\
K_4 &= 6.
\end{align*}

Each subregion $\domain_j$ is divided into $K_j$ tiles of equal measure, $\subd_{j,1}, \subd_{j,2}, ..., \subd_{j,K_j},$, as shown in Fig. \ref{fig:btsexample}. Then the agent performs SWEEP-SERVICE on the tiles in the following repeating sequence. During each phase, one tile from each subregion is swept:

\begin{align*}
\mathrm{phase}_1 &: \{ \subd_{1,1}, \subd_{2,1}, \subd_{3,1}, \subd_{4,1} \}, \\
\mathrm{phase}_2 &: \{ \subd_{1,1}, \subd_{2,2}, \subd_{3,2}, \subd_{4,2} \}, \\
\mathrm{phase}_3 &: \{ \subd_{1,1}, \subd_{2,1}, \subd_{3,3}, \subd_{4,3} \}, \\
\mathrm{phase}_4 &: \{ \subd_{1,1}, \subd_{2,2}, \subd_{3,1}, \subd_{4,4} \}, \\
\mathrm{phase}_5 &: \{ \subd_{1,1}, \subd_{2,1}, \subd_{3,2}, \subd_{4,5} \}, \\
\mathrm{phase}_6 &: \{ \subd_{1,1}, \subd_{2,2}, \subd_{3,3}, \subd_{4,6} \}. \\
\end{align*}

Again, the relative magnitudes of the areas of the subregions are irrelevant to the algorithm. The subregions in this example are of equal area only for clarity and simplicity. Of course, the absolute magnitudes of the areas relative to the agent's sensing radius will directly influence the system time achieved by the algorithm. Also, the subregions need not be convex as convexity is not required to compute an efficient path coverage. Furthermore, the subregions need not be connected. We only require that the time required to travel between them be negligible relative to the time required to perform path coverage sweeps on them. 

\begin{figure}[h]
\centering
\subfigure[Example environment and distribution.]{
   \includegraphics[scale = 0.3] {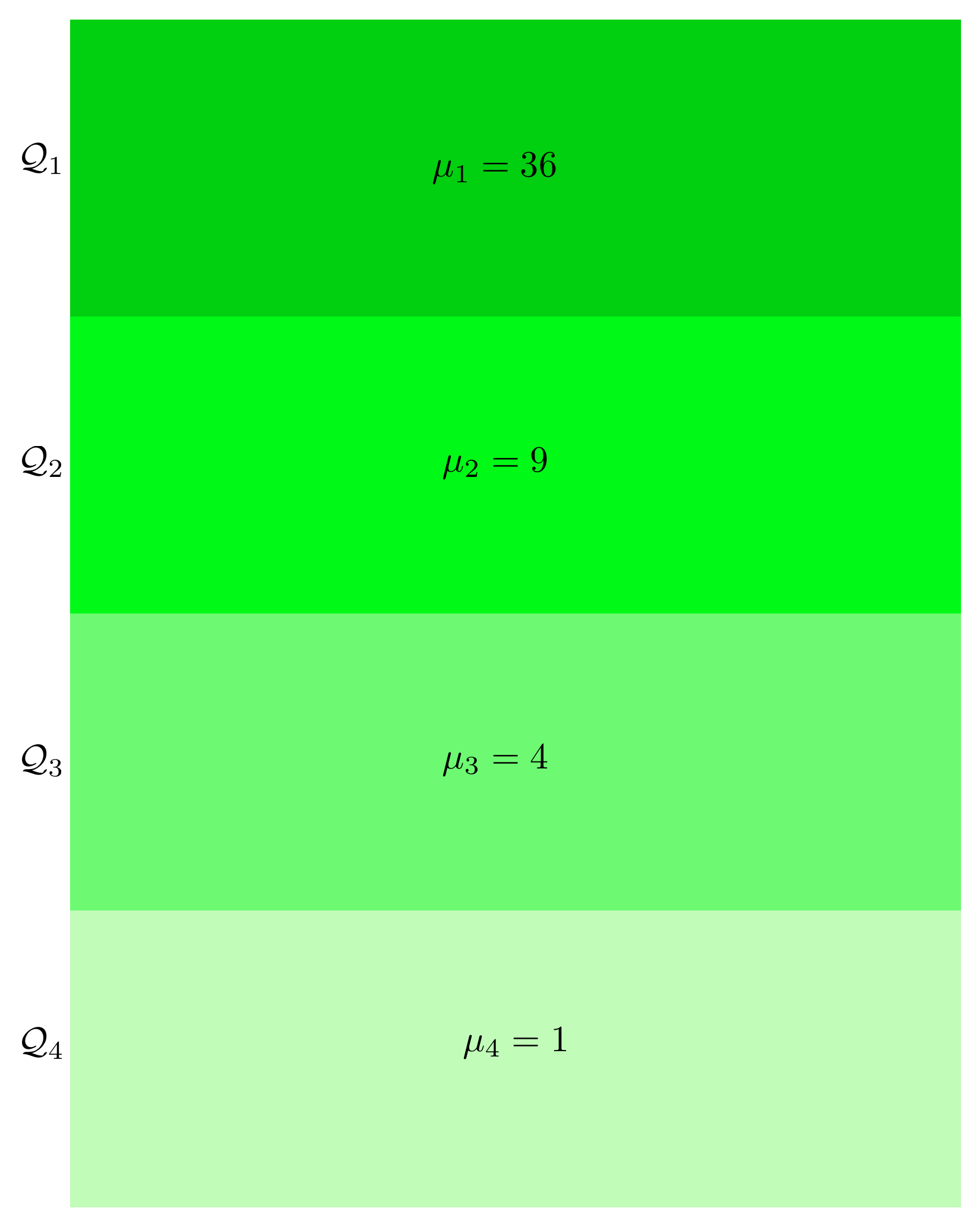}
   \label{fig:btsenv}
 }
 \subfigure[Tiling created by BTS policy.]{
   \includegraphics[scale = 0.3] {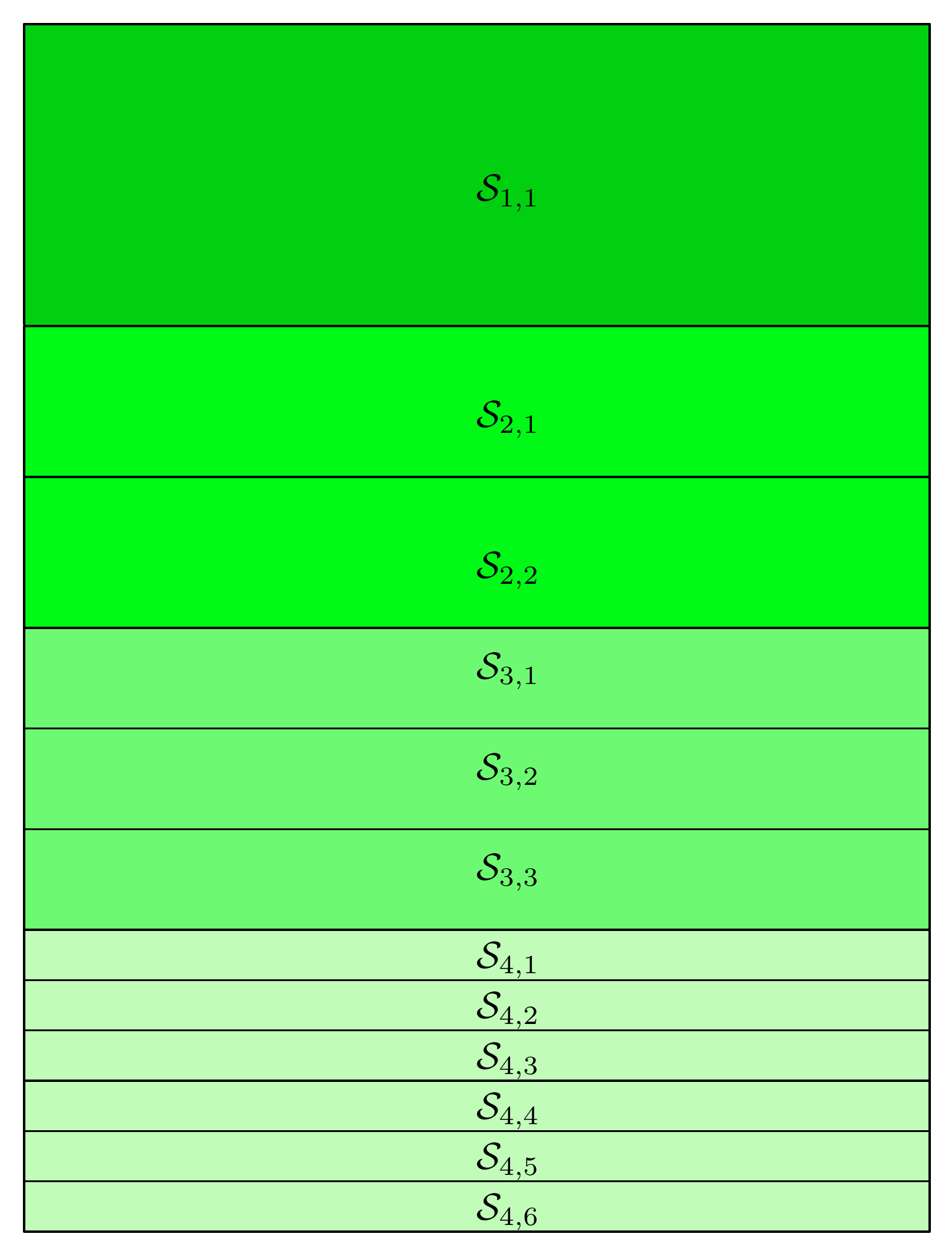}
   \label{fig:simbts}
 }

\label{fig:btsexample}
\caption{Example environment and resultant tiling created by BTS policy.}
\end{figure}

\end{example}

\begin{theorem}
\label{thm:bts}
If $\varphi$ is a piecewise uniform density and $\overline T_\mathrm{opt}$ is the optimal system time for the single-agent Limited Sensing DTRP over the class of spatially biased policies, then the system time of a single agent operating on $\domain$ under the BTS policy satisfies 
\begin{equation}
\frac{\overline T_\mathrm{BTS}}{\overline T_\mathrm{opt}} = 1 \ \ \ \ \mbox{as } r \rightarrow 0^+.
\end{equation}
\end{theorem}
\proof
The total distance traveled between tiles during a phase is no more than $J \diam(\domain)$.  Assuming that the policy is stabilizing, the number of targets serviced during the $i$-th phase $n_i$ is finite.  The expected length of the $i$-th phase $T^\mathrm{phase}_{i}$, conditioned on $n_{i}$, satisfies
$$ \E{T^\mathrm{phase}_{i} | n_{i}} \le  \frac{\sum_{j=1}^J L_\mathrm{swp}(\subd_{j,k_j})+ 2rn_{i}+J\diam(\domain)}{v}.$$
Applying Proposition \ref{thm:sweep},
\begin{align*}
\lim_{r\rightarrow 0^+} \E{T^\mathrm{phase}_{i} | n_{i}} r &\le \\
\lim_{r\rightarrow 0^+}  \frac{\sum_{j=1}^J L_\mathrm{swp}(\subd_{j,k_j}) r}{v} &+ \lim_{r\rightarrow 0^+}  \frac{2n_{i}r^2+J\diam(\domain)r}{v} \\
&=  \frac{\sum_{j=1}^J \area_{j}/K_j }{2v} \\
&=  \frac{\sum_{j=1}^J  \area_j \sqrt{  \mu_j}}{2Kv}. \\
\end{align*}
Conditioned upon its location $q \in \domain_j$, a target waits one half of $K_j$ phases to be serviced, 
\begin{align*}
\E{\overline T_\mathrm{BTS} | q \in \domain_j} &= \frac{1}{2}K_j T^\mathrm{phase} \\
&=  \frac{1}{2}\frac{K}{\sqrt{\mu_j}} T^\mathrm{phase}.
\end{align*}
Noting that $\mathrm{Pr}[q \in \domain_j] = \mu_j \area_j$ and unconditioning on $q \in \domain_j$ to find the system time,
\begin{align*}
\overline T_\mathrm{BTS}  &=\sum_{j=1}^J \mathrm{Pr}[q \in \domain_j] \cdot \E{\overline T_\mathrm{BTS} | q \in \domain_j} \\
&= \sum_{j=1}^J(  \area_j \mu_j )\cdot \left( \frac{1}{2}\frac{K}{\sqrt{\mu_j}} T^\mathrm{phase} \right)\\
&=\frac{K T^\mathrm{phase}}{2} \sum_{j=1}^J  \area_j \sqrt{\mu_j} .
\end{align*}
Thus,
\begin{align*}
\lim_{r\rightarrow 0^+} \overline T_\mathrm{BTS} r &= \left( \frac{K}{2} \sum_{j=1}^J  \area_j \sqrt{\mu_j} \right) \lim_{r \rightarrow 0^+}  T^\mathrm{phase} r \\
&=  \frac{1}{4v} \left( \sum_{j=1}^J  \area_j \sqrt{\mu_j}  \right)^2.
\end{align*}
If $\varphi$ is a piecewise uniform density and $m=1$, the lower bound on the optimal system time within the class of spatially biased policies in Theorem \ref{thm:sslb} takes on the form
\begin{equation*}
\lim_{r \rightarrow 0^+ }\overline T_\mathrm{opt} r \ge \frac{1}{4v}  \left( \sum_{j=1}^J \area_j \sqrt{\mu_j}  \right)^2.
\end{equation*}
Combining, the claim is proved.  
\endproof

Theorem \ref{thm:bts} shows that the optimal spatially biased policy for small sensor-range searches a point in the environment at regular intervals at a relative frequency proportional to $\sqrt{\varphi (q)}$, as was suggested by Eq. \eqref{eq:convexsoln} in the proof of the corresponding spatially biased lower bound in Theorem \ref{thm:sslb}.   

\begin{figure}[h]
\centering
\subfigure[Drawing of the environment and spatial distribution (parameterized using $\epsilon$) used for the numerical experiments on the BTS policy.]{
   \includegraphics[scale = 0.4] {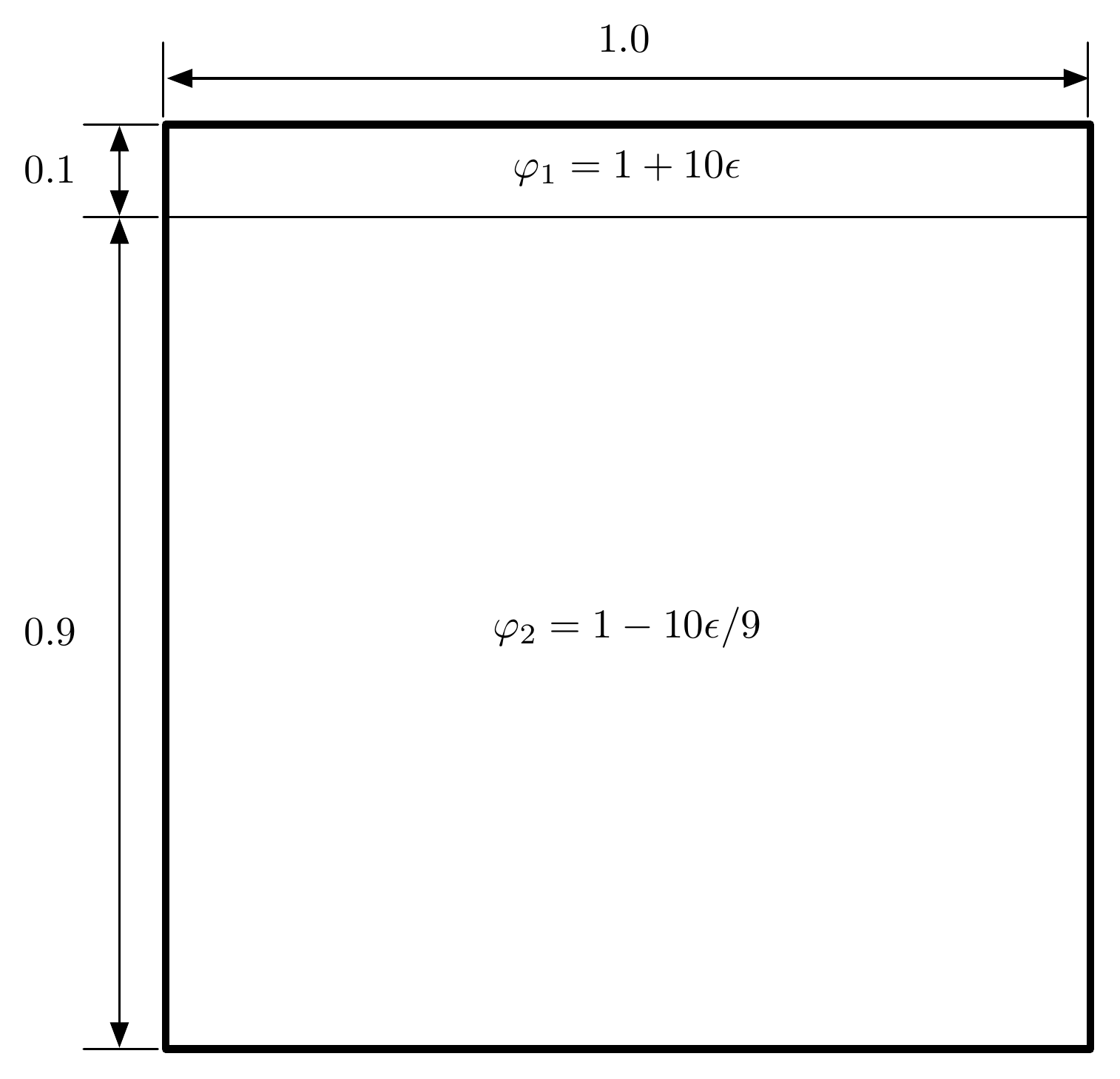}
   \label{fig:btsenv}
 }
 \subfigure[Simulation performance of the BTS policy for a single agent in a unit square environment with unit velocity and sensing radius $r=0.00625$. Results are compared with the theoretical lower bound (Eq. \ref{eq:ssblb}) for varying spatial distribution on a semi-log plot. The parameter $\epsilon$ was varied from $0$ to $0.89$, i.e., the spatial distribution varied from uniform to one in which $99\% $ of the incidents occuring in the subregion with $10\%$ of the area.]{
   \includegraphics[scale = 0.35] {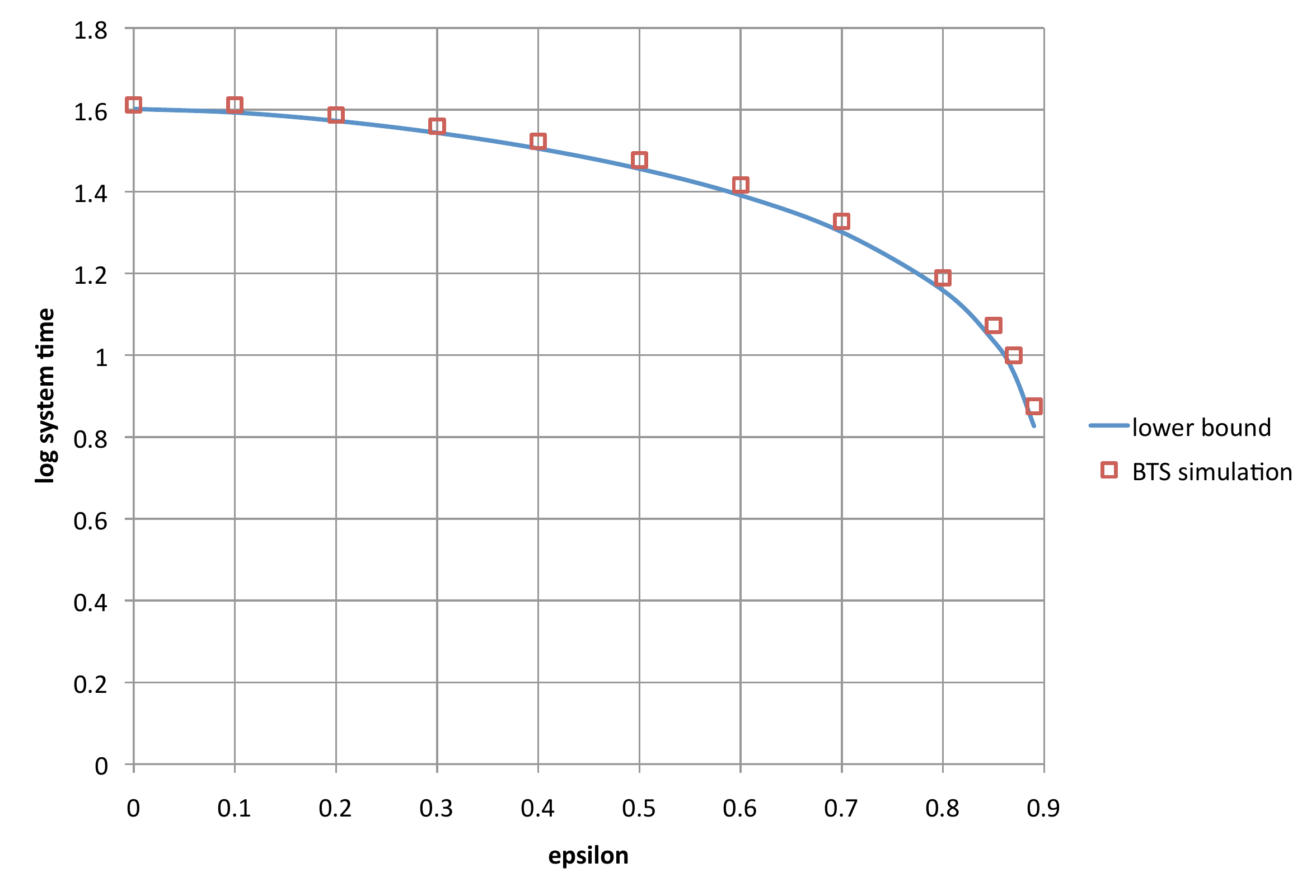}
   \label{fig:simbts}
 }

\label{fig:bts}
\caption{Numerical experiments of the BTS policy.}
\end{figure}

We performed numerical experiments of the BTS policy for a single agent. As shown in Fig. \ref{fig:bts} \subref{fig:btsenv}, the spatial distribution of the target-generation process was piece-wise uniform with a density of $\varphi_1 = 1+10\epsilon$ in the smaller region of area $0.1$ and $\varphi_2 = 1+10\epsilon/9$ in the larger region of area $0.9$. We varied $\epsilon$ from $0$ to $0.89$. This tested the algorithm's performance under a large range of spatial distributions: from uniform ($\epsilon=0$) to one in which targets appear in the smaller region with a $99\%$ probability ($\epsilon=0.89$). Results are shown in comparison with the lower bound (Eq. \ref{eq:ssblb}) in Fig. \ref{fig:bts} \subref{fig:simbts} on a semi-log plot. The BTS policy provides near-optimal performance for a large range of spatial distributions, i.e., the BTS policy adapts the distribution of the searching agent's position in order to exploit the spatially biased target-generation process and thereby reduce the expected system time overall. In other words, the searching agent provides higher quality of service to the targets in the higher density regions. Specifically, under the BTS policy, a target's expected quality of service will scale (relative to other targets in other regions) with the inverse square root of its region's density.

\subsection{The Unbiased Tile TSP (UTTSP) Policy}
This policy requires a tiling of the environment $\domain$ with the following properties.  For some positive integer $K \in \naturals$, partition $\domain$ into tiles $\subd_1,\subd_2,\dots,\subd_K$ such that 
$$\int_{\subd_k} \sqrt{\varphi(q)} dq = \frac{1}{K} \int_{\domain} \sqrt{\varphi(q)} dq, \ k = 1,2,\dots,K. $$  
In other words, it requires a convex and equitable partition of the environment $\domain$ with respect to the measure $\psi$ where $\psi (q) = \sqrt{\varphi (q)}$ for all $q \in \domain$.  Furthermore, the size and shape of each tile must be such that it can be contained in the sensor footprint of an agent, i.e., for each $\subd_k$ there exists a point $p_k$ such that $\|p_k-q\| \le r$ for all $q \in \subd_k$.  For example, if each tile can be bounded by a rectangle, neither of whose side-lengths exceeds $r/\sqrt{2}$, then this property is achieved.  A tiling with all these properties can be constructed with the following simple method.  First, partition $\domain$ into strip-like tiles of equal measure with $K_1$ parallel lines.  If $K_1$ is sufficiently large, then the width of the thinnest tile is less than or equal to $r/\sqrt{2}$.  Next, partition each of the $K_1$ strip-like tiles into $K_2$ tiles of equal measure using lines perpendicular to the first set.  If $K_2$ is sufficiently large, then the height of all tiles is less than or equal to $r/\sqrt{2}$.  The UTTSP policy is defined in Algorithm \ref{alg:uttsp}.  The index $i$ is a label for the current phase of the policy.      

\begin{algorithm} 
\caption{UTTSP Policy} 
\For{$i\leftarrow 1$ \KwTo $\infty$}{ 
\For{$k\leftarrow 1$ \KwTo $K$}{
\BlankLine
Execute SNAPSHOT-TSP on tile $\subd_{k}$
\BlankLine
} 
} 
\label{alg:uttsp} 
\end{algorithm}

\begin{theorem}
\label{thm:uttsp}
Let $\overline T_\mathrm{opt}$ be the optimal system time for the single-agent Limited Sensing DTRP over the class of spatially unbiased policies.  Then the system time of a single agent operating on $\domain$ under the UTTSP policy satisfies 
\begin{equation}
\frac{\overline T_\mathrm{UTTSP}}{\overline T_\mathrm{opt}} = 1 \ \ \ \ \mbox{as } \lambda \rightarrow \infty.
\end{equation}
\end{theorem}
\proof
Let us denote $T^\mathrm{tsp}_{k,i}$ as the time required to execute the tour of the targets in tile $\subd_k$ computed by SNAPSHOT-TSP in the $i$-th phase.  Including the time traveling between tiles, we note that 
$$T^\mathrm{phase}_i \le \sum_{k=1}^K T^\mathrm{tsp}_{k,i}+\frac{K \diam(\domain)}{v}.$$  
Applying Proposition \ref{thm:snap}, the time required for the first tile in the $(i+1)$-th phase $T^\mathrm{tsp}_{1,i+1}$, conditioned on the length of the $i$-th phase $T^\mathrm{phase}_i$, satisfies
\begin{align*}
\lim_{\lambda \rightarrow \infty} \frac{\E{T^\mathrm{tsp}_{1,i+1} | T^\mathrm{phase}_i}}{\sqrt{\lambda}} &= \frac{\beta \sqrt{T^\mathrm{phase}_i}}{v}  \int_{\subd_1}\sqrt{ \varphi (q)} \; dq \\
&= \frac{\beta \sqrt{T^\mathrm{phase}_i}}{Kv}  \int_{\domain}\sqrt{ \varphi (q)} \; dq.
\end{align*}
Since the choice of the first tile, and the epoch of the phase, is arbitrary, the expected time to visit 
all targets in a tile is uniform over all tiles. Summing over all tiles, we get
\begin{align*}
\lim_{\lambda \rightarrow \infty} \frac{\E{T^\mathrm{phase}_{i+1} | T^\mathrm{phase}_i}}{\sqrt{\lambda}} &= \\
K \cdot \lim_{\lambda \rightarrow \infty} \frac{\E{T^\mathrm{tsp}_{1,i+1} | T^\mathrm{phase}_i}}{\sqrt{\lambda}} &+\lim_{\lambda \rightarrow \infty} \frac{K \diam(\domain)}{v \sqrt{\lambda}} \\
&= \frac{\beta \sqrt{T^\mathrm{phase}_i}}{v}  \int_{\domain}\sqrt{ \varphi (q)} \; dq.
\end{align*}
From the equation above, it can be verified that
\begin{align*}
\E{T^\mathrm{phase}_{i+1} | T^\mathrm{phase}_i} &\ge T^\mathrm{phase}_i \\ 
\mbox{if} & \ \ \  T^\mathrm{phase}_i \le  \frac{\beta^2 \lambda}{v^2} \left( \int_{\domain}\sqrt{ \varphi (q)} \; dq \right)^2 
\end{align*}
and
\begin{align*}
\E{T^\mathrm{phase}_{i+1} | T^\mathrm{phase}_i} &\le T^\mathrm{phase}_i \\ 
\mbox{if} & \ \ \  T^\mathrm{phase}_i \ge  \frac{\beta^2 \lambda}{v^2} \left( \int_{\domain}\sqrt{ \varphi (q)} \; dq \right)^2 .
\end{align*}
Thus, the sequence of phases exhibits a fixed-point, from which
$$\lim_{i \rightarrow \infty}\E{T^\mathrm{phase}_i} = \frac{\beta^2 \lambda}{v^2} \left( \int_{\domain}\sqrt{ \varphi (q)} \; dq \right)^2.$$
Substituting, as $i \rightarrow \infty$,
\begin{align*}
\E{T^\mathrm{tsp}_{1,i}} &= \frac{\beta  \sqrt{\lambda} \sqrt{T^\mathrm{phase}_i}}{Kv}  \int_{\domain}\sqrt{ \varphi (q)} \; dq  \\
&=  \frac{\beta^2 \lambda}{Kv^2}  \left( \int_{\domain}\sqrt{ \varphi (q)} \; dq \right)^2 \\
& = \frac{T^\mathrm{phase}_i}{K} .
\end{align*}
Note that the above is quantity is independent of the specific tile.  In expectation, a target waits one half of a phase to enter a snapshot.  Because of the randomization in starting point and direction performed in SNAPSHOT-TSP, in expectation, a target waits one half of the time required to visit all targets in its snapshot.  Thus,
\begin{align*}
\E{\overline T_\mathrm{UTTSP}} &= \frac{1}{2} T^\mathrm{phase} + \frac{1}{2}\frac{T^\mathrm{phase}}{K} \\
&= \left( 1+\frac{1}{K} \right) \frac{\beta^2 \lambda}{2v^2} \left( \int_{\domain}\sqrt{ \varphi (q)} \; dq \right)^2.
\end{align*}
Since the system time is independent of the location of the target, the policy is spatially unbiased.  Using the optimal system time for the full-information DTRP in heavy load within the class of spatially unbiased policies in Eq. \eqref{eq:bertsimas} with $m=1$ as a lower bound, we see that for large $K$, the claim is proved. 
\endproof

Although the constraint on spatial bias does not allow the policy to service denser regions with higher frequency, Theorem \ref{thm:uttsp} shows that non-uniformity in the spatial distribution of targets, $\varphi$, still leads to a lowering of the optimal system time.  This is due to the efficiency gained by the ETSP tours due to non-uniformity, evident in Theorem \ref{thm:steele}.    

\subsection{The Biased Tile TSP (BTTSP) Policy} 
This policy requires that $\varphi$ be a piecewise uniform density, i.e., $\domain_1,\domain_2,\dots,\domain_J$ be a partition of $\domain$ such that $\varphi(q) = \mu_j \ \forall q \in \domain_j, j=1,2,\dots,J$.  Let us assume that each subset $\domain_j$ is convex, as a non-convex subset can be further partitioned into convex subsets.  Let us denote $\area_j = \mathrm{Area}(\domain_j), \ j=1,2,\dots,J.$  

This policy requires a tiling of the environment $\domain$ with the following properties.  For some positive integer $K \in \naturals$, partition each subset $\domain_j$ into $K_j = K/\mu_j^{1/3}$ convex tiles, each of area $\area_j/K_j = \area_j \mu_j^{1/3}/K.$  We assume $K$ is chosen large enough that an integer $K_j$ can be found such that $K/K_j$ is sufficiently close to $\mu_j^{1/3}.$  Furthermore, the size and shape of each tile must be such that it can be contained in the sensor footprint of an agent, i.e., for each $\subd_k$ there exists a point $p_k$ such that $\|p_k-q\| \le r$ for all $q \in \subd_k$.  The grid-like equitable tiling described for the UTTSP policy, applied here, would require that each $K_j$ be factorable into possibly large numbers $K_1$ and $K_2$.  This is undesirable because the numbers $K_j$ must maintain specific ratios related to the density $\mu_j$ in their domain $\domain_j$ as described above.  However, this scenario does have one simpler facet: the density functions within each $\domain_j$ are constant.   One example of a method for reaching convex and equitable partitions is given in~\cite{Pavone.Frazzoli.ea:CDC08}.   
Heuristically speaking, the proposed algorithms converge to configurations in which all cells are approximately hexagonal for constant measure $\psi$ and large $K_j$.  Hence, for sufficiently large $K$, these hexagonal tiles fit within a circle of radius $r$.  Let us give the tiles of $\domain_j$ an ordered labeling $\subd_{j,1},\subd_{j,2},\dots,\subd_{j,K_j}$.  The BTTSP policy is defined in Algorithm \ref{alg:bttsp}.  The index $i$ is a label for the current phase of the policy.    

\begin{algorithm} 
\caption{BTTSP Policy} 
Initialize $k_j \leftarrow 1$ for $j = 1,2,\dots,J$\\
\For{$i\leftarrow 1$ \KwTo $\infty$}{ 
\For{$j\leftarrow 1$ \KwTo $J$}{
Execute SNAPSHOT-TSP on tile $\subd_{j,k_j}$\\
\lIf{$k_j < K_j$}{$k_j \leftarrow (k_j+1)$}\\
\lElse{$k_j \leftarrow 1$}
} 
} 
\label{alg:bttsp} 
\end{algorithm}

\begin{theorem}
\label{thm:bttsp}
If $\varphi$ is a piecewise uniform density and $\overline T_\mathrm{opt}$ is the optimal system time for the single-agent Limited Sensing DTRP over the class of spatially biased policies, then the system time of a single agent operating on $\domain$ under the BTTSP policy satisfies 
\begin{equation}
\frac{\overline T_\mathrm{BTTSP}}{\overline T_\mathrm{opt}} = 1 \ \ \ \ \mbox{as } \lambda \rightarrow \infty.
\end{equation}
\end{theorem}
\proof
Assume, for now, that the sequence of phases exhibits a fixed-point, from which
$$\lim_{i \rightarrow \infty}\E{T^\mathrm{phase}_i} = T^\mathrm{phase}_\mathrm{ss},$$
where $T^\mathrm{phase}_\mathrm{ss}$ denotes the steady-state phase length.  Denote $T^\mathrm{tsp}_{j,k_j,i}$ as the time required to execute the tour of the targets in tile $\subd_{j,k_j}$ computed by SNAPSHOT-TSP in the $i$-th phase.  Any tile in $\domain_j$ waits $K_j$ phases between snapshots.  Applying Proposition \ref{thm:snap} with the fixed-point assumption, as $i \rightarrow \infty$,  
\begin{align*}
\lim_{\lambda \rightarrow \infty} \frac{\E{T^\mathrm{tsp}_{j,k_j,i}}}{\sqrt{\lambda}} &= \frac{\beta}{v} \sqrt{K_jT^\mathrm{phase}_\mathrm{ss}}  \int_{\subd_{j,k_j}}\sqrt{ \varphi (q)} \; dq \\
&= \frac{\beta}{v}  \sqrt{K_jT^\mathrm{phase}_\mathrm{ss}} \frac{\area_j}{K_j}\sqrt{\mu_j} \\
&=  \frac{\beta}{v}  \sqrt{\frac{T^\mathrm{phase}_\mathrm{ss}}{K}} \area_j \mu_j^{2/3}. 
\end{align*}
In other words, the fixed-point assumption implies that
\begin{align*}
\lim_{i \rightarrow \infty} \E{T^\mathrm{tsp}_{j,k_j,i}}  &= T^\mathrm{tsp}_{j,k_j,\mathrm{ss}} =  \frac{\beta}{v}  \sqrt{\frac{\lambda T^\mathrm{phase}_\mathrm{ss}}{K}} \area_j \mu_j^{2/3}, 
\end{align*}
for $k_j = 1,2,\dots,K_j.$
Summing over all tiles in a phase, and including distance traveled between tiles,
\begin{align*}
\lim_{\lambda \rightarrow \infty} \frac{\E{T^\mathrm{phase}_\mathrm{ss}}}{\sqrt{\lambda}} &= \sum_{j=1}^J \lim_{\lambda \rightarrow \infty} \frac{ \E{T^\mathrm{tsp}_{j,k_j,\mathrm{ss}}}}{\sqrt{\lambda}} +  \lim_{\lambda \rightarrow \infty} \frac{J \cdot \diam (\domain)}{v \sqrt{\lambda}} \\ 
&= \frac{\beta}{v}  \sqrt{\frac{ T^\mathrm{phase}_\mathrm{ss}}{K}} \sum_{j=1}^J\area_j \mu_j^{2/3}.
\end{align*}
The above implies that
$$T^\mathrm{phase}_\mathrm{ss} = \frac{\beta^2 \lambda}{K v^2} \left( \sum_{j=1}^J\area_j \mu_j^{2/3} \right)^2.$$
We now investigate the expected system time of a target, conditioned upon its location $q \in \domain_j$.  The time a target waits before entering a snapshot, $T_\mathrm{snap}^-$, is one half of $K_j$ phases, i.e.,
\begin{align*}
\E{T_\mathrm{snap}^- | q \in \domain_j}&= \frac{1}{2} K_j T^\mathrm{phase}_\mathrm{ss} \\
&= \frac{1}{2}K T^\mathrm{phase}_\mathrm{ss} \mu_j^{-1/3} .
\end{align*}
Unconditioning on the location of the target,
\begin{align*}
\E{T_\mathrm{snap}^- }&= \sum_{j=1}^J \mathrm{Pr} \left[ q \in \domain_j \right] \cdot \E{T_\mathrm{snap}^- | q \in \domain_j} \\
&= \sum_{j=1}^J \left(\area_j \mu_j \right) \cdot \left(\frac{1}{2}K T^\mathrm{phase}_\mathrm{ss}   \mu_j^{-1/3} \right) \\
&= \frac{1}{2}K T^\mathrm{phase}_\mathrm{ss} \sum_{j=1}^J \area_j \mu_j^{2/3} \\
&= \frac{\beta^2 \lambda}{2 v^2} \left( \sum_{j=1}^J\area_j \mu_j^{2/3} \right)^3.
\end{align*}
Because of the randomization in starting point and direction performed in SNAPSHOT-TSP, in expectation, the time a target waits after entering a snapshot, $T_\mathrm{snap}^+$, is one half of the time required to visit all targets in its snapshot.  
\begin{align*}
\E{T_\mathrm{snap}^+ | q \in \domain_j}&= \frac{1}{2}  T^\mathrm{tsp}_{j,k_j,\mathrm{ss}} \\
&= \frac{1}{2} \frac{\beta}{v}  \sqrt{\frac{\lambda T^\mathrm{phase}_\mathrm{ss}}{K}} \area_j \mu_j^{2/3} .
\end{align*}
Unconditioning on the location of the target,
\begin{align*}
\E{T_\mathrm{snap}^+ }&= \sum_{j=1}^J \mathrm{Pr} \left[ q \in \domain_j \right] \cdot \E{T_\mathrm{snap}^+ | q \in \domain_j} \\
&= \sum_{j=1}^J \left(\area_j \mu_j \right) \cdot \left( \frac{1}{2} \frac{\beta}{v}  \sqrt{\frac{\lambda T^\mathrm{phase}_\mathrm{ss}}{K}} \area_j \mu_j^{2/3} \right) \\
&=  \frac{\beta}{2v}  \sqrt{\frac{\lambda T^\mathrm{phase}_\mathrm{ss}}{K}} \sum_{j=1}^J \area_j^2 \mu_j^{5/3} \\
&= \frac{\beta^2 \lambda}{2 v^2} \frac{1}{K} \left( \sum_{j=1}^J\area_j \mu_j^{2/3} \right)\left( \sum_{j=1}^J \area_j^2 \mu_j^{5/3} \right).
\end{align*}
Combining,
\begin{align*}
\overline T_\mathrm{BTTSP} &= \E{T_\mathrm{snap}^- }+\E{T_\mathrm{snap}^+ }\\
&= \frac{\beta^2 \lambda}{2 v^2}   \left( \sum_{j=1}^J\area_j \mu_j^{2/3} \right)^3 \\ 
&+ \frac{1}{K}  \frac{\beta^2 \lambda}{2 v^2} \left( \sum_{j=1}^J\area_j \mu_j^{2/3} \right)\left( \sum_{j=1}^J \area_j^2 \mu_j^{5/3} \right) .
\end{align*}
Namely, for large $K$, we have
\begin{align*}
\lim_{\lambda \rightarrow \infty} \frac{\overline T_\mathrm{BTTSP} }{\lambda} = \frac{\beta^2 }{2 v^2}  \left( \sum_{j=1}^J\area_j \mu_j^{2/3} \right)^3.
\end{align*}
If $\varphi$ is a piecewise uniform density and $m=1$, then the optimal system time for the full-information DTRP in heavy load within the class of spatially biased policies in Eq. \eqref{eq:bertsimas2} takes on the form
\begin{equation}
\lim_{\lambda \rightarrow \infty} \frac{\overline T_\mathrm{opt}}{\lambda} = \frac{\beta^2}{2v^2} \left( \sum_{j=1}^J \area_j \mu_j^{2/3}  \right)^{3}.
\end{equation}
Using the above as a lower bound for the Limited Sensor DTRP in heavy load, we see that for large $K$, the claim is proved.  
\endproof

Theorem \ref{thm:bttsp} shows that the optimal spatially biased policy for heavy load services targets in the vicinity of a point $q$ at regular intervals at a relative frequency proportional to $\varphi (q)^{1/3}$.  

\section{Multiple Agents with Equitable Regions of Dominance}
\label{sec:multiple}

We have presented four algorithms and proven them optimal in four respective cases of the single-vehicle Limited Sensing DTRP.  We wish to adapt these policies to the multiple-vehicle scenario, retaining optimality, with minimal communication and  collaboration among the agents.  Towards this end, consider the following strategy.  Given a single-vehicle policy $\pi$, partition the environment $\domain$ into regions of dominance $\vdom(\domain) = (\vdom_1,\vdom_2,...,\vdom_m)$ where $\cup_{\ell=1}^m \vdom_\ell = \domain$ and $\vdom_\ell \cap \vdom_h = \emptyset$ if $\ell \ne h$.  Each vehicle executes policy $\pi$ on its own region of dominance.  In other words, agent $\ell$ is responsible for all targets appearing in $\vdom_\ell$ and ignores all others.  

The control policies to reach convex and equitable partitions proposed in~\cite{Pavone.Frazzoli.ea:CDC08} are distributed and can be performed through $m$ mobile agents, where each agent only need communicate with agents in neighboring cells of the partition.  We do not pursue this facet of the multi-agent system further, as these methods are available and well studied.  In the following theorems, we show that if the regions of dominance are a convex and equitable partition of $\domain$ with respect to a designed measure $\psi$, then the decentralized strategy achieves an optimal system time.  However, the appropriate measure $\psi$ depends on the problem parameters (small sensing radius or heavy load) and spatial constraints (biased or unbiased).  Moreover, it is a function of the spatial distribution of targets $\varphi$ in three of the four cases.       

Some of the single-vehicle policies in the previous section require the knowledge of certain properties of the environment such as its target generation rate, and its spatial target distribution.  A single-vehicle policy operating on a subregion $\vdom_\ell \subseteq \domain$ takes as input the local target-generation process of $\vdom_\ell$ with time intensity $\lambda_\ell = \lambda \varphi(\vdom_\ell)$ and spatial distribution $\map{\varphi_\ell}{\vdom_\ell}{\reals_+}$ where $\varphi_\ell(q) = \varphi(q)/\varphi(\vdom_\ell)$ if $q \in \vdom_\ell$ and $\varphi_\ell(q) = 0$ otherwise.  Note that $\varphi_\ell$ is normalized such that $\varphi_\ell(\vdom_\ell) = 1.$  

\begin{theorem}
\label{thm:murs}
Let $\overline T_\mathrm{opt}$ be the optimal system time for the $m$-agent Limited Sensing DTRP over the class of spatially unbiased policies, and let $\vdom(Q)$ be a convex and equitable partition of $\domain$ with respect to measure $\psi$ where $\psi(q) = 1$ for all $q \in \domain$.  If each agent $\ell \in I_m$ operates on its own region of dominance $\vdom_\ell$ under the URS policy, then the system time satisfies
\begin{equation}
\frac{\overline T_\mathrm{URS/ERD}}{\overline T_\mathrm{opt}} = 1 \ \ \ \ \mbox{as } r \rightarrow 0^+.
\end{equation}
\end{theorem}
\proof 
Denoting $\area_\ell = \mathrm{Area}(\vdom_\ell)$, we apply Theorem \ref{thm:urs} to find the system time of a target, conditioned upon its location,
\begin{align*}
\E{T_\mathrm{URS/ERD} | q \in \vdom_\ell} = \frac{\area_\ell}{4vr}.
\end{align*}
Since $\psi(\vdom_\ell) = \area_\ell$, the equitable partition implies that $\area_\ell = \area/m$ for all $\ell \in I_m$.  Hence
\begin{align*}
\E{T_\mathrm{URS/ERD} | q \in \vdom_\ell} = \frac{\area}{4mvr},
\end{align*}
and since the above is independent of the targets location, it is in fact the unconditioned system time.  Moreover this implies that the strategy is spatially unbiased.  Combining with Theorem \ref{thm:sslb}, the claim is proved.  
\endproof

\begin{theorem}
\label{thm:mbts}
Let $\varphi$ be a piecewise uniform density, let $\overline T_\mathrm{opt}$ be the optimal system time for the $m$-agent Limited Sensing DTRP over the class of spatially biased policies, and let $\vdom(Q)$ be a convex and equitable partition of $\domain$ with respect to measure $\psi$ where $\psi(q) = \sqrt{\varphi(q)}$ for all $q \in \domain$.  If each agent $\ell \in I_m$ operates on its own region of dominance $\vdom_\ell$ under the BTS policy, then the system time satisfies
\begin{equation}
\frac{\overline T_\mathrm{BTS/ERD}}{\overline T_\mathrm{opt}} = 1 \ \ \ \ \mbox{as } r \rightarrow 0^+.
\end{equation}
\end{theorem}
\proof 
We apply Theorem \ref{thm:bts} to find the system time of a target, conditioned upon its location,
\begin{align*}
\E{T_\mathrm{BTS/ERD} | q \in \vdom_\ell} = \frac{1}{4vr}  \left( \int_{\vdom_\ell} \sqrt{\varphi_\ell(q)} \ dq \right)^2.
\end{align*}
Substituting $\varphi_\ell(q) = \varphi(q)/\varphi(\vdom_\ell)$ we get
\begin{align*}
\E{T_\mathrm{BTS/ERD} | q \in \vdom_\ell} = \frac{1}{4vr} \frac{1}{\varphi(\vdom_\ell)}  \left( \int_{\vdom_\ell} \sqrt{\varphi(q)} \ dq \right)^2.
\end{align*}
Unconditioning on the location of the target,
\begin{align*}
\overline T_\mathrm{BTS/ERD} &= \sum_{\ell = 1}^m \mathrm{Pr} \left[q \in \vdom_\ell \right] \cdot \E{T_\mathrm{BTS/ERD} | q \in \vdom_\ell} \\ 
&= \sum_{\ell = 1}^m \varphi(\vdom_\ell) \cdot \frac{1}{4vr} \frac{1}{\varphi(\vdom_\ell)}  \left( \int_{\vdom_\ell} \sqrt{\varphi(q)} \ dq \right)^2 \\
&= \frac{1}{4vr} \sum_{\ell = 1}^m  \left( \int_{\vdom_\ell} \sqrt{\varphi(q)} \ dq \right)^2.
\end{align*}
But the equitable partition with respect to $\psi(q) = \sqrt{\varphi(q)}$ implies that 
\begin{align*}
  \int_{\vdom_\ell} \sqrt{\varphi(q)} \ dq =   \frac{1}{m} \int_{\domain} \sqrt{\varphi(q)} \ dq, \ \ \forall \ell \in I_m,
\end{align*}
and so
\begin{align*}
\overline T_\mathrm{BTS/ERD} = \frac{1}{4mvr}  \left( \int_{\domain} \sqrt{\varphi(q)} \ dq \right)^2.
\end{align*}
Combining with Theorem \ref{thm:sslb}, the claim is proved.  
\endproof

\begin{theorem}
\label{thm:muttsp}
Let $\overline T_\mathrm{opt}$ be the optimal system time for the $m$-agent Limited Sensing DTRP over the class of spatially unbiased policies, and let $\vdom(Q)$ be a convex and equitable partition of $\domain$ with respect to measure $\psi$ where $\psi(q) = \sqrt{\varphi(q)}$ for all $q \in \domain$.  If each agent $\ell \in I_m$ operates on its own region of dominance $\vdom_\ell$ under the UTTSP policy, then the system time satisfies
\begin{equation}
\frac{\overline T_\mathrm{UTTSP/ERD}}{\overline T_\mathrm{opt}} = 1 \ \ \ \ \mbox{as } \lambda \rightarrow \infty.
\end{equation}
\end{theorem}
\proof 
We apply Theorem \ref{thm:uttsp} to find the system time of a target, conditioned upon its location,
\begin{align*}
\E{T_\mathrm{UTTSP/ERD} | q \in \vdom_\ell} = \frac{\beta^2}{2v^2} \lambda_\ell  \left( \int_{\vdom_\ell} \sqrt{\varphi_\ell(q)} \ dq \right)^2.
\end{align*}
Substituting $\varphi_\ell(q) = \varphi(q)/\varphi(\vdom_\ell)$ and $\lambda_\ell = \lambda \varphi(\vdom_\ell)$ we get
\begin{align*}
\E{T_\mathrm{UTTSP/ERD} | q \in \vdom_\ell} = \frac{\beta^2 \lambda}{2v^2}   \left( \int_{\vdom_\ell} \sqrt{\varphi(q)} \ dq \right)^2.
\end{align*}
But the equitable partition with respect to $\psi(q) = \sqrt{\varphi(q)}$ implies that 
\begin{align*}
  \int_{\vdom_\ell} \sqrt{\varphi(q)} \ dq =   \frac{1}{m} \int_{\domain} \sqrt{\varphi(q)} \ dq, \ \ \forall \ell \in I_m,
\end{align*}
and so
\begin{align*}
\E{T_\mathrm{UTTSP/ERD} | q \in \vdom_\ell} = \frac{\beta^2 \lambda}{2m^2v^2}   \left( \int_{\domain} \sqrt{\varphi(q)} \ dq \right)^2.
\end{align*}
Since the above is independent of the targets location, it is in fact the unconditioned system time.  Moreover this implies that the strategy is spatially unbiased.  Combining with Eq. \eqref{eq:bertsimas}, the claim is proved.
\endproof

\begin{theorem}
\label{thm:mbttsp}
Let $\varphi$ be a piecewise uniform density, let $\overline T_\mathrm{opt}$ be the optimal system time for the $m$-agent Limited Sensing DTRP over the class of spatially biased policies, and let $\vdom(Q)$ be a convex and equitable partition of $\domain$ with respect to measure $\psi$ where $\psi(q) = \varphi(q)^{2/3}$ for all $q \in \domain$.  If each agent $\ell \in I_m$ operates on its own region of dominance $\vdom_\ell$ under the BTTSP policy, then the system time satisfies
\begin{equation}
\frac{\overline T_\mathrm{BTTSP/ERD}}{\overline T_\mathrm{opt}} = 1 \ \ \ \ \mbox{as } \lambda \rightarrow \infty.
\end{equation}
\end{theorem}
\proof 
We apply Theorem \ref{thm:bttsp} to find the system time of a target, conditioned upon its location,
\begin{align*}
\E{T_\mathrm{BTTSP/ERD} | q \in \vdom_\ell} = \frac{\beta^2}{2v^2} \lambda_\ell  \left( \int_{\vdom_\ell} \varphi_\ell(q)^{2/3} \ dq \right)^3.
\end{align*}
Substituting $\varphi_\ell(q) = \varphi(q)/\varphi(\vdom_\ell)$ and $\lambda_\ell = \lambda \varphi(\vdom_\ell)$ we get
\begin{align*}
\E{T_\mathrm{BTTSP/ERD} | q \in \vdom_\ell} = \frac{\beta^2 \lambda}{2v^2}  \frac{1}{\varphi(\vdom_\ell)} \left( \int_{\vdom_\ell} \varphi(q)^{2/3} \ dq \right)^3.
\end{align*}
Unconditioning on the location of the target,
\begin{align*}
\overline T_\mathrm{BTTSP/ERD} &= \sum_{\ell = 1}^m \mathrm{Pr} \left[q \in \vdom_\ell \right] \cdot \E{T_\mathrm{BTTSP/ERD} | q \in \vdom_\ell} \\ 
&= \sum_{\ell = 1}^m \varphi(\vdom_\ell) \cdot \frac{\beta^2 \lambda}{2v^2}  \frac{1}{\varphi(\vdom_\ell)} \left( \int_{\vdom_\ell} \varphi(q)^{2/3} \ dq \right)^3 \\
&=  \frac{\beta^2 \lambda}{2v^2}  \sum_{\ell = 1}^m \left( \int_{\vdom_\ell} \varphi(q)^{2/3} \ dq \right)^3.
\end{align*}
But the equitable partition with respect to $\psi(q) = \varphi(q)^{2/3}$ implies that 
\begin{align*}
  \int_{\vdom_\ell}\varphi(q)^{2/3} \ dq =   \frac{1}{m} \int_{\domain} \varphi(q)^{2/3} \ dq, \ \ \forall \ell \in I_m,
\end{align*}
and so
\begin{align*}
\overline T_\mathrm{BTTSP/ERD} = \frac{\beta^2 \lambda}{2m^2v^2}   \left( \int_{\domain} \varphi(q)^{2/3} \ dq \right)^3.
\end{align*}
Combining with Eq. \eqref{eq:bertsimas2}, the claim is proved.
\endproof

In summary, we have offered four policies, each of which performs optimally for the four cases studied: small sensing-range (spatially biased and unbiased) and heavy load (spatially biased and unbiased).  In addition, we have offered a method by which to adapt the four policies to a multi-vehicle setting, retaining optimality, with minimal communication or collaboration.  In particular, the agents partition the environment into regions of dominance, and each vehicle executes the single-vehicle policy on its own region, ignoring all others.  However, the nature of the partition varies in the different cases addressed. Each scenario requires regions of dominance equitable with respect to a measure appearing in the optimal system time of the corresponding single-vehicle case.  In the spatially unbiased small sensing-range case, the regions of dominance are equitable with respect to area.  Interestingly, the spatially biased small sensing-range case, and the spatially unbiased heavy load case both require regions of dominance equitable with respect to measure $\psi$ where $\psi(q) = \sqrt{\varphi (q)}$ for all $q \in \domain$.  Finally, the spatially biased heavy load case requires regions of dominance equitable with respect to measure $\psi$ where $\psi(q) = \varphi (q)^{2/3}$ for all $q \in \domain$.   

We have shown that in heavy load, the limited information gathering capabilities of the agents have no effect on their achievable performance.  We suggest the intuitive explanation that in heavy load, the environment is dense with targets, and so the searching component added to the DTRP is non-existent.  On the other hand, for small sensing range, the lack of information gathering capability detracts from the achievable performance of the system.  However, the presented policies for these cases were still proved optimal through new lower bounds related to the searching capability of the agents and the necessary structure of any stabilizing policy.  

\section{Conclusions}
\label{sec:con}
We have addressed a multi-agent problem with information constraints we call the DTRP with limited sensing. Our analysis yields precise characterizations of the system time, and the parameters describing the capabilities and limitations of the agents and the nature of the environment appear (or don't appear) in these expressions, giving insight into how the parameters affect (or don't affect) the efficiency of the system. We summarize the characterizations of the optimal system times for the four cases studied in Table \ref{table:summary}.

\begin{table}
\label{table:summary}
  \caption{Optimal system time for the DTRP with limited sensing.}
\resizebox{8cm}{!} {
  \begin{tabular}{| c | c | c | }
    \hline 
     & spatially unbiased & spatially biased \\ \hline
   $r \rightarrow 0^+$  & $\frac{\mathrm{A}}{4mvr}$ & $ \frac{1}{4mvr}  ( \int \sqrt{\varphi} \ dq )^2$ \\ \hline
     $\lambda \rightarrow \infty$  & $ \frac{\beta^2 \lambda^2}{2m^2v^2} ( \int \sqrt{\varphi} \ dq )^2 $& $ \frac{\beta^2 \lambda^2}{2m^2v^2} ( \int \varphi^{2/3} \ dq )^3$\\
    \hline
  \end{tabular}
}
\end{table}

In terms of methods and approaches, we note the path taken in the small-sensor case.  In order to place lower bounds on the achievable performance of any algorithm, we carefully relax constraints to arrive at a convex optimization problem whose solution offers insight into the structure of the optimal algorithm. Using this structure as guidance, we design a provably optimal algorithm.  Moreover, we have made use of results from the mature fields of combinatorial optimization and probability theory.  

Another interesting result is that the limited sensing capability has no impact on performance when the target arrival-rate is high.  In other words, this lack of global information does not hinder an agent from efficient routing choices.  This result assumes knowledge of prior statistics on the global environment.  Perhaps through proper mechanism design, a game-theoretic approach~\cite{Arslan.Marden.ea:07,Shamma:07} might integrate the learning of the global structure of the environment and the adaptation of policy choices.  Consider a team of agents with limited sensing and no communication~\cite{Arsie.Savla.Frazzoli:TAC09} operating in a common environment.  From the perspective of an individual agent, there might not be any observed difference between i) a region with low target-arrival rate, and ii) a region with high target-arrival rate that is frequently serviced by other agents.  Does this difference matter?  In either case, the agent should search for target-rich regions in order to maximize both its own utility (target-servicing rate) and the value it is adding to the multi-agent system.

{\small \bibliographystyle{ieeetr}%
  \bibliography{enright}}

\end{document}